\documentclass{egpubl}

\usepackage[T1]{fontenc}
\usepackage{dfadobe}  

\usepackage{cite}  
\BibtexOrBiblatex

\electronicVersion
\PrintedOrElectronic
\ifpdf \usepackage[pdftex]{graphicx} \pdfcompresslevel=9
\else \usepackage[dvips]{graphicx} \fi

\usepackage{egweblnk} 

\usepackage{lipsum}
\usepackage{times}
\usepackage{epsfig}
\usepackage{amsmath}
\usepackage{amssymb}
\usepackage{subcaption}
\usepackage{stfloats}
\usepackage{mathtools}
\usepackage{amsfonts}
\usepackage{ragged2e}
\usepackage{comment}
\usepackage{booktabs, multirow, tabularx}
\newcolumntype{L}{>{\RaggedRight}X}
\usepackage{setspace}
\usepackage[ruled,vlined]{algorithm2e} 
\SetKwComment{Comment}{ $\triangleright$ \ }{}
\SetCommentSty{textrm}
\SetKwFor{Iterate}{$\forall$}{}{}
\SetKwFor{If}{$\textrm{if} \ $}{}{}
\usepackage{caption}
\usepackage[hang,flushmargin]{footmisc}
\usepackage{cprotect}
\usepackage{xfrac}

\newcommand{\SO}[1]{\ensuremath{\textrm{SO}(#1)}}
\newcommand{\SE}[1]{\ensuremath{\textrm{SE}(#1)}}

\newcommand{\norm}[1]{\ensuremath{\left \Vert #1 \right \Vert}}
\newcommand{\abs}[1]{\ensuremath{\left \vert #1 \right \vert}}

\newcommand{\vn}[1]{\ensuremath{{\mathbf #1}}}

\DeclarePairedDelimiterX{\normSized}[1]{\lVert}{\rVert}{#1}

\newcommand{\ignore}[1]{}
\newcommand{\T}{{\mathfrak{ T}}}

\title[Efficient Spatially Adaptive Convolution and Correlation]{Efficient Spatially Adaptive Convolution and Correlation}

\author[T. W. Mitchel et al.]
{\parbox{\textwidth}{\centering Thomas W. Mitchel \quad Benedict Brown  \quad David Koller \quad Tim Weyrich \quad Szymon Rusinkiewicz \quad Michael Kazhdan}
}

\begin{document}

\maketitle
\begin{abstract}
Fast methods for convolution and correlation underlie a variety of
applications in computer vision and graphics, including efficient
filtering, analysis, and simulation.  However, standard convolution and
correlation are inherently limited to fixed filters:  spatial adaptation is
impossible without sacrificing efficient computation.  In early work,
Freeman and Adelson~\shortcite{Freeman:PAMI:1991} have shown how steerable
filters can address this limitation, providing a way for rotating the
filter as it is passed over the signal.

In this work, we provide a general, representation-theoretic, framework
that allows for spatially varying linear transformations to be applied to
the filter.  This framework allows for efficient implementation of extended
convolution and correlation for transformation groups such as rotation (in
2D and 3D) and scale, and provides a new interpretation for previous
methods including steerable filters and the generalized Hough transform. 
We present applications to pattern matching, image feature description, vector field visualization,
and adaptive image filtering.

\printccsdesc   
\end{abstract}
 

\begin{figure*}[t]
\epsfig{file=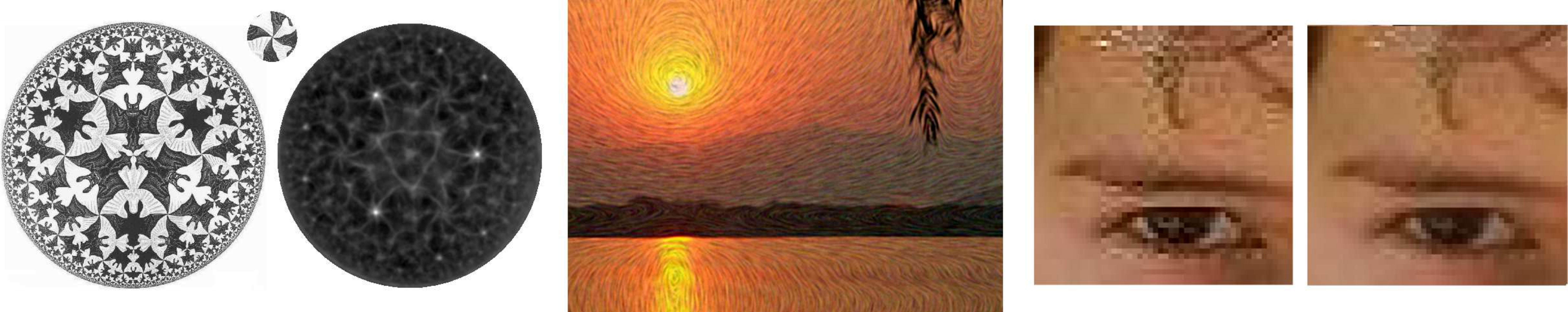,width=2\columnwidth}
\caption{\footnotesize
\label{fig:teaser}
Applications of extended convolution.
\textbf{Left:}  Rotation-independent pattern matching was used to locate the
pattern in the image at left.  The three correct matches correspond to
the three peaks in the match-quality image.
\textbf{Center:}  A rotation-dependent filter applied to a photograph with
added noise produces an artistic effect.
\textbf{Right:} Scale-dependent smoothing is used to remove compression
artifacts from an image while preserving edges.
}
\end{figure*}

\section{Introduction}
\label{s:intro}
One of the most widely used results in signal processing is the
convolution theorem, which states that convolution in the spatial domain
is equivalent to multiplication in the frequency domain.
Combined with the availability of Fast Fourier Transform
algorithms~\cite{Cooley:MC:1965,Frigo:IEEE:2005}, it reduces the
complexity of what would be a quadratic-time operation to a nearly
linear-time computation (linearithmic).  This, together with the closely-related correlation
theorem, have enabled efficient
algorithms for applications in many domains, including audio analysis and
synthesis~\cite{Allen:ASSP:1977,Moorer:IEEE:1977}, 
pattern recognition and compression of
images~\cite{Kumar:2005,Wallace:CACM:1991}, symmetry detection in 2D
images~\cite{Keller:TIP:2006} and 3D models~\cite{Kazhdan:SGP:2004},
reconstruction of 3D surfaces~\cite{Schall:GMP:2006}, inversion of the
Radon transform for medical imaging~\cite{Kak:2001,Naterrer:SIAM:2001},
and solving partial differential~\cite{Iorio:2001} and fluid
dynamic equations~\cite{Kass:SIGGRAPH:1990,Stam:JGT:2001}.

Despite the pervasiveness of convolution in signal processing, it has an
inherent limitation: when convolving a signal with a filter, the filter
remains fixed throughout the convolution, and cannot adapt to spatial
information.

\paragraph*{Early Work on Spatially-Varying Filters}
A simple approach to allowing spatial variation is to limit the number of
different filters that are allowed.  For example, if differently-rotated
versions of a filter are required, it is possible to quantize the rotation
angle, compute a (relatively) small number of standard convolutions, and
select the closest-matching rotation at each pixel.

Motivated by the early research of Knutsson {\em et al.} on non-stationary
anisotropic filtering~\shortcite{Knutsson:TOC:1983}, Freeman and
Adelson~\shortcite{Freeman:PAMI:1991} investigated the idea of
\emph{steerable filters}. The essential observation is that, for angularly
band-limited filters, the results of spatially adaptive filtering with
\emph{arbitrary} per-pixel rotation can be computed from per-pixel linear
combinations of a finite set of convolutions with rotated filters. 
Given appropriate conditions on the filter, different transformation groups
can be accommodated in this
framework~\cite{Freeman:PAMI:1991,Simoncelli:IP:1996,Teo:TPAMI:1999}.

\paragraph*{Contribution}
In this work, we provide a representation-theoretic interpretation of
steerable filters, and explore generalizations enabled by this
interpretation.  Our key idea is to focus not on the properties of filters
that allow ``steerability,'' but rather on the \emph{structure of the
group} from which transformations are drawn.  Specifically, we show
how the ability to perform efficient function steering is related to the
decomposition of the space of filters into irreducible representations of
the transformation group.  The analysis permits us to answer key questions such
as:
\begin{itemize}
\item How many convolutions are required for spatially-adaptive filtering,
given a particular transformation group and a particular filter?
\item Given a fixed budget of convolutions, what is the best way to approximate
the spatially-adaptive filtering?
\end{itemize}
We are able to answer these questions not only for 2D rotation, but for a
variety of transformations including scaling and non-commutative groups
such as 3D rotation.  Moreover, we show that it is possible to obtain
significantly better approximation results than previous methods that attempt
to discretize the space of transformations.

One of our main generalizations is to apply our results to both the
convolution and correlation operations, for which the effect of a spatially
varying filter has different natural interpretations.  Extended convolution
is naturally interpreted as a \emph{scattering} operation, in which the
influence of each point in the signal is distributed according to the
transformed filter.  In contrast, extended correlation has a natural
interpretation as a \emph{gathering} operation, in which each output pixel
is the linear combination of input pixels weighted by the
locally-transformed filter.  We show that the two approaches are
appropriate for different applications, in particular demonstrating that
the spatially adaptive voting of the Generalized Hough
Transform~\cite{Ballard:PR:1981} may be implemented using extended
convolution with a specially-designed filter.

Figure~\ref{fig:teaser} shows several applications of spatially adaptive
filtering that are enabled using our extended correlation and convolution. 
At left, we show pattern matching that locates all rotated instances of the
pattern (top) in the target image (far left).  At center we demonstrate an
image manipulation in which gradient directions are used to place
anisotropic brushstrokes across the image.  At right, we show the effect of
denoising an image with a filter whose scale is controlled by gradient
magnitude, which yields edge-preserving smoothing similar to anisotropic
diffusion and bilateral filtering~\cite{Perona:PAMI:1990,Weickert:SSTCV:1997,Tomasi:ICCV:1998,Slate:TOG:2006}.
For the first two applications, we use spatially adaptive scattering to
detect the pattern and to distribute the brushstrokes, respectively.  For
the third, we use our generalization of function steering to support
spatially adaptive (data-dependent) scaling of a smoothing filter. 

\paragraph*{Approach} Our approach is to leverage the fact that
linear transformations of the filter can be realized as invertible matrices
acting on a high-dimensional vector space (the space of functions,
corresponding to filters). Choosing a basis for the space of functions,
the transformation associated to a spatial location can be expressed in
matrix form and spatially adaptive filtering can be implemented as a sum
of standard convolutions over the matrix entries (Section~\ref{s:approach}).
When the basis is chosen to conform to the irreducible representations
of the transformation group, the matrix becomes block-diagonal with repeating
diagonal blocks (Section~\ref{s:representation}), thereby reducing the total
number of convolutions that need to be performed.

The generality of the method makes it capable of supporting a number of
image processing and image analysis operations. In this paper, we highlight
its versatility by describing several of these applications, including the use
of the extended convolution in two different types of pattern matching applications
(Sections~\ref{s:matching}, \ref{s:evaluation}, and \ref{s:contour}) and  three different types of image filtering applications
(Section~\ref{s:smoothing}). Additionally, we provide a discussion of how filter
steering can be generalized to three dimensions, where the group of rotations is no
longer commutative (Section~\ref{s:steering3D}).

\section{Defining Adaptive Filtering}
\label{s:definition}
We begin by formalizing our definitions of spatially adaptive filtering. Following
the nomenclature of stationary signal processing, we consider both the correlation
and convolution of a signal with a filter. Though these operators are identical
in the stationary case, up to reflection of the filter through the origin, they define
different notions of spatially adaptive filtering.

For both we assume that we are given a spatial function $H$, a filter $F$, and a
transformation field $\T$ that defines how the filter is to be transformed at each
point in the domain.

\subsection{Correlation}
The correlation of $H$ with $F$ is defined at a point $p$ as:
$$\bigl(H\star F\bigr)(p) = \int H(q) \, \overline{F(q-p)}\,dq.$$
Using the notation $\rho_p$ to denote the operator that translates functions by the
vector $p$:
$$\bigl(\rho_pF\bigr)(q) \equiv F(q-p),$$
we obtain an expression for the correlation as:
$$\bigl(H\star F\bigr)(p) = \int H(q)\,\overline{\bigl(\rho_pF\bigr)(q)}\,dq.$$
That is, the correlation of $H$ with $F$ can be thought of as a {\em gathering}
operation in which the value at a point $p$ is defined by first translating $F$
to the point $p$ and then accumulating the values of the conjugate of $F$,
weighted by the values of $H$.

We generalize this to spatially adaptive filtering, defining the value of the
{\em extended correlation} at a point $p$ as the value obtained by first applying
the transformation $\T(p)$ to the filter, translating the transformed filter
to $p$, and then accumulating the conjugated values of the transformed
$F$, weighted by $H$:
\begin{equation}
\{H,\ \T\} \star F = \int H(q)\,\overline{\rho_p\bigl(\T(p)\,F\bigr)(q)} \, dq.
\label{eq:Def-ECorr}
\end{equation}
Note that, if the transformations $\T$ are linear, extended correlation
maintains standard correlation's properties of being linear in the signal
and conjugate-linear in the filter.

\subsection{Convolution}
Similarly, convolution can be expressed as:
$$\bigl(H*F\bigr)(p) = \int H(q)\,\bigl(\rho_qF\bigr)(p)\,dq.$$
In this case, the convolution of $H$ with $F$ can be thought of as a {\em scattering}
operation, defined by iterating over all points $q$ in the domain, translating $F$ to
each point $q$, and distributing the values of $F$ weighted by the value of $H(q)$.

Again, we generalize this to spatially adaptive filtering, defining the {\em extended
convolution} by iterating over all points $q$ in the domain, applying the transformation
$\T(q)$ to the filter $F$, translating the transformed filter to $q$, and then
distributing the values of the transformed $F$, weighted by $H(q)$:
\begin{equation}
\{H, \T\} * F = \int H(q)\,\rho_q\bigl(\T(q)F\bigr)(p) \, dq.
\label{eq:Def-EConv}
\end{equation}
Note that, as with standard convolution, the extended convolution is linear
in both the signal and the filter (if the transformations $\T$ are linear).

\subsection{A Theoretical Distinction}
While similar in the case of stationary filters, these operators give rise to different
types of image processing techniques in the context of spatially adaptive filtering. This
distinction becomes evident if we consider the response of filtering a signal that is the
delta function centered at the origin.

In the case of extended convolution, the response is the function
$\T(0)\,F$, corresponding to the transformation of the filter by the
transformation defined at the origin.  In the case of the extended
correlation, the response is more complicated: the value at point
$p$ comes from the conjugate of the filter at the point(s)
$\bigl(\T(p)\bigr)^{-1}(p)$.  Since the transformation field is changing,
this implies that some of the values of the filter can be represented by
multiple positions in the response, while others might not be represented
at all.

Beyond thinking of ``gathering'' and ``scattering,'' another way of
understanding the distinction between how correlation and convolution
extend to varying filters is by considering the dependency of the
transformation on the variables.  In extended correlation, the filter's
transformation depends on the spatial variable of the result.  In contrast,
in extended convolution the transformation depends on the variable of
integration.  This provides another way of deciding which of
the operations will be appropriate for a given problem.

\subsection{A Practical Distinction}
\label{s:distinction}
The distinction between the two is also evidenced in a more practical setting, if one compares
using steerable filters~\cite{Freeman:PAMI:1991} with using the Generalized Hough
Transform~\cite{Ballard:PR:1981} for pattern detection where a local frame can be
assigned to each point.

\paragraph*{Steerable Filters} For steerable filters, pattern detection is
performed using extended correlation. The filter corresponds to an aligned pattern template,
and detection is performed by iterating over the pixels in the image, aligning the filter to
the local frame, and gathering the correlation into the pixel. The pixel with the largest
correlation value is then identified as the pattern center.

\paragraph*{Generalized Hough Transform} For the Generalized Hough Transform, 
pattern detection is performed using extended convolution. The filter corresponds to candidate
offsets for the pattern center and detection is performed by iterating over the pixels in the
image, aligning the filter to the local frame, and distributing votes into the candidate
centers, weighted by the confidence that the current pixel lies on the pattern. The pixel
receiving the largest number of votes is then identified as the pattern center.

\section{A First Pass at Adaptive Filtering}
\label{s:approach}
In this section, we show that for linear transformations, extended correlations
and convolutions can be performed by summing the results of several standard
convolutions. If we do not restrict the space of possible transformations, little
simplification is possible (either mathematically or in algorithm design) to the
brute-force computation implied by Equations~\ref{eq:Def-ECorr} and~\ref{eq:Def-EConv}.
Therefore, we restrict our filter functions to lie within an
$n$-dimensional space $\mathcal F$, spanned by (possibly complex-valued)
basis functions $\langle F_1,\ldots,F_n\rangle$.  Moreover, we restrict
the transformations $\T(p):  \mathcal F \rightarrow \mathcal F$ to act
linearly on functions, meaning that they can be represented with matrices
(possibly with complex entries).  This permits significant simplification.

We expand the filter as $F=[F_1\,\ldots\,F_n]\,[a_1\,\ldots\,\,a_n]^T$, and
write each transformation $\T(p)$ as a matrix with entries $\T_{ij}(p)$. 
Thus we can express the transformation of $F$ by $\T(p)$ as the linear
combination:
$$
\T(p)\,F = \T(p)\!\left(\sum_{i=1}^n a_i F_i\right)
 = \sum_{i,j=1}^n \T_{ij}(p)\,a_j F_i.
$$
This, in turn, gives an expression for extended correlation as:
\begin{eqnarray}
\bigl(\{H,\ \T\} \star F\bigr)(p)
&=&\int H(q)\,
\overline{\rho_p\!\left(\sum_{i,j=1}^n \T_{ij}(p)\,a_jF_i\right)\!(q)}
\,dq\nonumber\\
&=&\sum_{i,j=1}^n \overline{\T_{ij}(p)}\int H(q)\,\overline{\rho_p\bigl(a_j F_i\bigr)(q)}\,dq\nonumber\\
\{H, \T\} \star F
&=&\sum_{i,j=1}^n\overline{\T_{ij}}\cdot\left( H \star a_jF_i\right),
\label{eq:generalECorr}
\end{eqnarray}
which can be obtained by taking the linear combination of standard correlations.
Similarly, we get an expression for extended convolution as:
\begin{equation}
\{H,\ \T\} * F =\sum_{i,j=1}^n \left(\T_{ij}\cdot H\right)*a_j F_i
\label{eq:generalEConv}
\end{equation}
which can also be obtained by taking the linear combination of standard convolutions.

Note that both equations can be further simplified to reduce the total number of standard
correlations (resp. convolutions) by leveraging the linearity of the correlation
(resp. convolution) operator:
\begin{align}
\label{eq:generalECorr2}
\{H, \ \T\} \star F &= \sum_{i=1}^n \left[\sum_{j=1}^n\overline{\T_{ij}a_j}\right]\cdot\left( H \star F_i\right)\\
\label{eq:generalEConv2}
\{H, \ \T\} *     F &= \sum_{i=1}^n \left(\left[\sum_{i=1}^n\T_{ij}a_j\right]\cdot H\right)*F_i.
\end{align}
However, we prefer the notation of Equations~\ref{eq:generalECorr} and~\ref{eq:generalEConv}
as they keep the filter separate from the signal, facilitating the discussion in the next section.

\subsubsection*{Example $n = 1$}
As a simple example, we consider the case in which we would like to correlate a signal
with an adaptively rotating filter $F$ which is supported within the unit disk and has values:
$$F(r,\theta) = a e^{ik\theta}.$$

In this case, rotating by an angle $\Theta$ amounts to multiplying the filter by
$e^{-ik\Theta}$. Thus, the extended correlation at point $p$ can be computed by
multiplying the filter $F$ by $e^{-ik \Theta(p)}$, where $\Theta(p)$ is the angle of rotation
at $p$, and then evaluating the correlation with the transformed filter at $p$. However,
since correlation is conjugate-linear in the filter, the value of the extended correlation can also
be obtained by first performing a correlation of $H$ with the un-transformed $F$, and only
then multiplying the result at point $p$ by $e^{ik\Theta(p)}$.

\subsubsection*{Example $n = 3$}
Next, we consider a more complicated example in which the filter $F$ resides within a three-dimensional
space of functions, $F = a_1 F_1 + a_2 F_2 + a_3 F_3$, with the basis defined as:
$$F_1(r,\theta)=e^{i2\theta},\quad F_2(r,\theta) = e^{-i2\theta},\quad F_3(r,\theta) = 1$$
In this case, rotating by an angle $\Theta$ amounts to multiplying the first component of the
filter by $e^{-i2\Theta}$, the second by $e^{i2\Theta}$, and the third by $1$ so the previous
approach will not work. However, by linearity, the extended correlation with $F$ can be expressed as the sum of the separate extended
correlations with $aF_1$, $bF_2$, and $cF_3$. Each of these can each be obtained by computing
the standard correlations with $a_1F_1$, $a_2F_2$, and $a_3F_3$ and then multiplying the values
at point $p$ by $e^{i2\Theta(p)}$, $e^{-i2\Theta(p)}$, and $1$ respectively. Thus, we can obtain
the extended correlation by performing $n=3$ separate correlations and taking their linear
combination.

With respect to the notation in Equation~\ref{eq:generalECorr}, rotating the filter $F$ by
an angle of $\Theta$ multiplies the coefficients $(a_1,a_2,a_3)^T$ by:
$$
\T_{ij}(\Theta)=
\left(\begin{array}{ccc}
e^{-i2\Theta} & 0 & 0\\
0 & e^{i2\Theta} & 0 \\
0 & 0 & 1
\end{array}\right).
$$
Thus, the extended correlation with $F$ can be computed by computing the standard correlations with
the $n^2=9$ functions $a_iF_j$, multiplying the results of these correlations by the functions $\T_{ij}(p)$,
and then taking the sum. However, since the functions $\T_{ij}(p)$ are uniformly zero whenever $i\neq j$,
the standard correlations with $a_iF_j$ become unnecessary for $i\neq j$, and the extended correlation
can be expressed using only $n=3$ standard correlations.

\subsubsection*{Example $n=3$, revisited}
Though the previous example shows that the extended correlation with $F$ can be computed efficiently,
we now show that the efficiency is tied to the way in which we factored the filter. In particular, we
show that if the wrong factorization is chosen, the cost of computing the extended correlation can
increase. Consider the same filter as above, but now expressed as the linear combination of a different
basis as $F(r,\theta) = \tilde{a}_1\tilde{F}_1(r,\theta) + \tilde{a}_2\tilde{F}_2(r,\theta) + \tilde{a}_3\tilde{F}_3(r,\theta)$,
with:
$$\tilde{F}_1(r,\theta)=\cos^2\!\theta,
~~
\tilde{F}_2(r,\theta) = \sin^2\!\theta,
~~
\tilde{F}_3(r,\theta) = \cos\theta\sin\theta.$$

Rotating such a filter by an angle of $\Theta$ multiplies the coefficients
$(\tilde{a}_1,\tilde{a}_2,\tilde{a}_3)^T$ by:
$$
\tilde{\T}_{ij}(\Theta)=
\left(\begin{array}{ccc}
\cos^2\Theta & \phantom{-}\sin^2\Theta & -\cos\Theta\sin\Theta\\
\sin^2\Theta & \phantom{-}\cos^2\Theta & \phantom{-}\cos\Theta\sin\Theta\\
\sin2\Theta  & -\sin2\Theta  & \cos2\Theta
\end{array}\right).
$$
Thus, the extended correlation with $F$ can be computed by computing the standard correlations with
the functions $\tilde{a}_i\tilde{F}_j$, multiplying the results of these correlations by the functions
$\tilde{\T}_{ij}(p)$ respectively, and then taking the sum. In this case, since the matrix entries are
all non-zero, all $n^2=9$ standard correlations are required.

Of course, the above discussion was purely a strawman:  using the grouping
of terms in Equations~\ref{eq:generalECorr2} and~\ref{eq:generalEConv2}, it
is possible to avoid the need for $n^2$ correlations.  However, focusing on
the structure of the $\T$ matrix and using the tools of representation
theory to find a basis in which it has a particularly simple structure, we
can bring the computational requirements even below $O(n)$ correlations or
convolutions.

\section{Choosing a Basis}
\label{s:representation}

As hinted at in the previous section, the efficiency of the implementation
of extended correlation (resp. convolution) is tied to
the choice of basis. In this section we make this explicit by showing that by choosing
the basis of functions appropriately, we obtain matrices that are sparse (with many zero
entries) and have repeated elements. Each zero and repetition corresponds to a standard
correlation (resp. convolution) that does not need to be computed.

We begin by considering the group of planar rotations. We show that there exists a basis of
functions in which the transformation matrix $\T$ becomes sparse, specifically diagonal.
We then use results from representation theory to generalize this, and
to establish limits on how sparse the matrix $\T$ can be made. We conclude this section with a
detailed discussion of the relation of our work to earlier work in steerable functions.

\subsection{Rotations}
\label{sec:rotation}
\def\sqrtn2{\ensuremath{\sqrt{n} \,\times \!\!\sqrt{n}}}
To motivate the result that the choice of basis is significant to the
structure of the matrix $\T$, consider planar rotations and
their effect on 2D functions.  In this case, the structure of $\T$ is
most easily exposed by considering the filter in polar coordinates.  In
particular, rotations preserve radius:  $(r,\theta)$ is necessarily
mapped to $(r,\theta')$.  Thus, in polar coordinates the only nonzero
entries in $\T$ occur in blocks around the diagonal, one block for each
$r$. Starting with an $n$-pixel image, transformation into polar
coordinates will give a function sampled at $N=O(n^{1/2})$ radii and
$K=O(n^{1/2})$ angles. Hence the nonzero entries in $\T$ will occupy
$N$ blocks of size $K\times K$.

To make $\T$ even more sparse, we consider representing the functions
at each radius in the frequency domain, rather than the spatial domain.
That is, within a radius the basis functions are proportional to
$e^{ik\theta}$, for a fixed $k$.  Applying a rotation to such a
single-frequency function preserves that frequency; it is, in fact,
expressible by multiplying the function by $e^{-ik\Theta}$,
where $\Theta$ is the angle of the rotation.  Therefore, in this basis
$\T$ has been simplified to purely diagonal (with complex entries).

In moving from an arbitrary basis to polar and polar/Fourier bases, we
have reduced the number of nonzero entries in $\T$ from $(N\times K)^2=O(n^2)$
to $N\times K^2=O(n^{1.5})$ to $N\times K=O(n)$. Correspondingly, the
number of standard correlations (resp. convolutions) that need to be
computed is also reduced.

There is one more reduction we may obtain by considering
\emph{repeated} entries in $\T$.  In particular, we observe that all the
diagonal entries corresponding to a particular frequency $k$, across different
radii, will be the same: $e^{-ik\Theta(p)}$. Although the rotation
angle $\Theta(p)$ may vary across the image, all of these entries
will vary in lock-step, and the associated diagonal entries $\T_{ii}$ will
be identical. Thus, we may perform all such correlations (resp. convolutions)
at once by correlating (resp. convolving) $e^{-ik\Theta(q)}$ with the sum of
all $a_i F_i$, where $F_i$ has angular frequency $k$. As a result, the number
of distinct correlations (resp. convolutions) is reduced to $K=O(n^{1/2})$.

Summarizing, to compute the extended correlation of a 2D signal $H$ and
rotation field $\T$ with filter $F$:

\paragraph*{Filter Decomposition}
We first decompose $F$ as the sum of functions with differing angular frequencies:
$$
F = \sum_{k=-K/2}^{K/2} F_k\qquad\hbox{with}\qquad F_k(r,\theta)=f_k(r)\,e^{ik\theta}.
$$
This can be done, for example, by first expressing $F$ in polar coordinates,
and then running the 1D FFT at each radius to get the different frequency coefficients.\hfill [$O(n+n\log n)$]

\paragraph*{Standard Correlation}
Next, we compute the standard correlations of the signal with the functions $F_k(r,\theta)=f_k(r)e^{ik\theta}$:
$$G_k = H \star F_k\qquad\hbox{for each }k\in[-K/2,K/2].$$
This can be done by first evaluating the function $f_k(r)e^{ik\theta}$ on a regular grid and
then using the 2D Fast Fourier Transform to perform the correlation. \hfill [$O(n^{3/2}+n^{3/2}\log n)$]

\paragraph*{Linear Combination}
Finally, we take the linear combination of the correlation results:
$$\bigl(\{H,\ \T\}\star F\bigr)(p) = \sum_{k=-K/2}^{K/2} e^{ik\Theta(p)}\,G_k(p),$$
weighting the contribution of the $k$-th correlation to the pixel $p$ by the
conjugate of the
$k$-th power of the unit complex number corresponding to the rotation at $p$.\hfill[$O(n^{3/2})$]

The extended convolution can be implemented similarly, but in this case we need
to pre-multiply the signal:
$$G_k(p) = H(p)\cdot e^{-ik\Theta(p)}\qquad\hbox{for each }k\in[-K/2,K/2]$$
and only then sum the convolutions of $G_k$ with $F_k$.

\subsection{Generalization}
In implementing the extended correlation  for rotations we have
taken advantage of the fact that the space of filters could be expressed
as the direct-sum of subspaces that (1)~are fixed under rotation, and
(2)~could be grouped into subspaces on which rotations act in a similar manner.

The decomposition of a space of functions into such subspaces is a
central task of representation theory, which tells us that {\em any}
vector-space $V$, acted upon by a group $G$, can be decomposed into a sum
of subspaces (e.g.~\cite{Serre:1977:SV}):
$$
V\cong \,\bigoplus_{\lambda}\, m_\lambda \! V^{\lambda},
$$
where $\lambda$ is the frequency, indexing the subspace fixed under the action of the group, and
$m_\lambda$ is the multiplicity of the subspace. While we are only guaranteed
that the subspace $V^\lambda$ is one-dimensional when the group $G$ is
commutative, the subspace $V^\lambda$ is guaranteed to be as small as possible
(i.e. irreducible) so that $V^\lambda$ cannot be decomposed further into
subspaces fixed under the action of $G$.

Using the decomposition theorem, we know that if ${\mathcal F}$ represents the
space of filters and the transformations $\T(p)$ belong to a group $G$, then
we can decompose ${\mathcal F}$ into irreducible representations of $G$:
\begin{equation}
{\mathcal F} = \,\bigoplus_{k=1}^{\gamma}
    \left(\bigoplus_{l=1}^{m_k}{\mathcal F}_{kl}\right)
\label{eq:decomposition}
\end{equation}
where $k$ indexes the sub-representation and, for a fixed $k$, the sub-representations
$\left\{{\mathcal F}_{kl}\right\}_{l=1}^{m_k}$ are all isomorphic.

Referring back to the discussion of rotation in Section~\ref{sec:rotation}, the group
acting on the filters is $G=\SO{2}$ (the group of rotations in the plane) and the
sub-representations ${\mathcal F}_{kl}$ are just functions of constant radius and
constant angular frequency.

\subsubsection{Block-Diagonal Matrix}
Using the decomposition in Equation~\ref{eq:decomposition}, we can
choose a basis for ${\mathcal F}$ by choosing a basis for each subspace
${\mathcal F}_{kl}$.
Since for fixed $k$ the $\{{\mathcal F}_{kl}\}_{l=1}^{m_k}$ are all isomorphic,
we can denote their dimension by $n_k$ and represent the basis for ${\mathcal F}_{kl}$ by:
$${\mathcal F}_{kl}=\hbox{Span}\langle F_1^{kl},\ldots,F_{n_k}^{kl}\rangle.$$

Additionally, since ${\mathcal F}_{kl}$ is a sub-representation, we know that
$\T(q)$ maps ${\mathcal F}_{kl}$ back into itself. This implies that
we can represent the restriction of $\T(q)$ to ${\mathcal F}_{kl}$ by an $n_k\times n_k$ matrix
with $(i,j)$-th entry $\T_{ij}^{kl}(q)$. Thus, given $F=\sum a_i^{kl}F_i^{kl}\in{\mathcal F}$, we can
express the transformation of $F$ by $\T(q)$ as:
\def\fudge{\vrule width 0pt height 3.7pt depth 0pt}
$$
\T(q)(F)=\!\sum_{k=1}^{{\gamma}_{\fudge}}\sum_{l=1}^{\,m_k}\sum_{i,j=1}^{n_k}
 \T_{ij}^{kl}(q)\,a_i^{kl}F_j^{kl}
$$
corresponding to a block-diagonal representation of $\T$ by a matrix with
$\sum m_k$ blocks, where the $(k,l)$-th block is of size $n_k\times n_k$.
As before, this gives:
$$
\{H, \ \T\}\star F =\!\sum_{k=1}^{{\gamma}_{\fudge}}\sum_{l=1}^{\,m_k}\sum_{i,j=1}^{n_k}
\overline{\T_{ij}^{kl}}\cdot\left(H \star a_i^{kl}F_j^{kl}\right).
$$

Using this decomposition, evaluating the extended correlation now requires
the computation of $m_1 n_1^2+\cdots+m_R\,n_R^2$ standard correlations. Note that, since
$n=m_1 n_1+\cdots+m_R\, n_R$, the number of linear combinations will be smaller
than $n^2$ if the space ${\mathcal F}$ contains more than one irreducible representation.

\subsubsection{Multiplicity of Representations}
We further improve the efficiency of the extended correlation by using the
multiplicity of the representations. Since the spaces $\{{\mathcal F}_{kl}\}_{l=1}^{m_k}$
correspond to the same representation, we can choose bases for them such that the matrix
entries $\T_{ij}^{kl}(q)$ have the property that $\T_{ij}^{kl}(q)=\T_{ij}^{kl'}(q) \equiv \T_{ij}^k$
for all $1\leq l,l'\leq m_k$. As a result, the extended correlation simplifies to:
$$
\{H, \ \T\}\star F=\!\sum_{k=1}^{{\gamma}_{\fudge}}\sum_{i,j=1}^{n_k}
  \Biggl[\overline{\T_{ij}^k}\cdot\left(H\star\biggl[\sum_{l=1}^{\,m_k}a_i^{kl}F_j^{kl}\!\biggr]\right)\Biggr].
$$

Thus, we only need to perform one standard correlation for each set of
isomorphic representations, further reducing the number of standard
correlations to $n_1^2+\cdots+n_R^2$.

While the previous discussion has focused on the extended correlation, an
analogous argument shows that the same decomposition of the space of filters results
in an implementation of the extended convolution that requires $n_1^2+\cdots+n_R^2$
standard convolutions.

\subsection{Band-Limiting}
In practice, we approximate the extended correlation (resp. convolution)
by only summing the contribution from the first $K\ll{\gamma}$ frequencies, for some constant $K$.
This further reduces the number of standard correlations (resp. convolutions) to
$n_1^2+\cdots+n_K^2$ and is equivalent to band-limiting the filter prior to the
computation of the extended convolution. For example, when rotating images sampled
on a regular grid with $n$ pixels, this can reduce the complexity of extended correlation
to $O(K\,n\log n)$ by band-limiting the filter's angular components.

\subsection{Relation to Steerable Filters}
Using the extended correlation, the method described above can be used
to perform efficient steerable filtering. While the implementation
differs from the one described in~\cite{Freeman:PAMI:1991}, the complexity
is identical, with both implementations running in $O(K\,N^2\log N)$ time
for $N\times N$ images and filters with maximal angular frequency~$K$.

We briefly review Freeman and Adelson's implementation of steerable
filtering and discuss how it fits into our representation-theoretic
framework. We defer the discussion of the limitations of the earlier
implementation in the context of higher-dimensional steering to
Section~\ref{s:steering3D}.

\label{ss:ClassicalSteering}
In the traditional implementation of steerable filters, the filter $F$ is
used to define the steering basis.  (Note that the original work of
Freeman and Adelson~\shortcite{Freeman:PAMI:1991} also proposes, but
does not use, an interpretation based on alternative basis functions.)
Specifically, when the filter is
angularly band-limited with frequency $K$, the steerable filtering is
performed using the functions $F_0,\ldots,F_{K-1}$, where the function
$F_k$ is the rotation of $F$ by an angle of $k\pi/K$.

Because the span of these functions is closed under rotation and because
it contains the filter $F$, the functions $F_0,\ldots,F_{K-1}$ can be used
for performing the extended correlation. In particular, one can compute
the matrix $\T_{ij}(\Theta)$ describing how the rotation of a basis
function can be expressed as a linear combination of the basis, and then
take the linear combinations of the standard correlations of the signal with
the functions $a_jF_i$ weighted by the matrix entries~$\T_{ij}$.

While this interpretation of steerable filtering within the context of our
representation-theoretic framework hints at an implementation requiring
$K^2$ standard correlations (since the entries $\T_{ij}$ are non-zero) this
is not actually the case. What makes the classical implementation of steerable
filtering efficient is that the filter is one of the basis vectors, $F=F_0$, so
the decomposition of the filter $F$ as $F=a_0F_0+\cdots+a_{K-1}F_{K-1}$, has
$a_0=1$ and $a_i=0$ for all $i\neq0$. Thus, while all $K^2$ matrix entries
$\T_{ij}$ are non-zero, only $K$ of the functions $a_jF_i$ are non-zero, so
the steerable filtering only requires that $K$ standard correlations be
performed.

\section{Application to Pattern Detection}
\label{s:matching}
We apply extended convolution to detect instances of a pattern within an
image, even if the pattern occurs at different orientations.  Recall that
this approach may be thought of as an instance of the generalized Hough
transform, such that image pixels \emph{vote} for locations consistent with
the presence of the pattern.  Figure~\ref{fig:teaser}, left, shows an
example application in which we search for instances of a pattern in
Escher's \emph{Heaven and Hell}.  In this example, all three rotated
versions of the pattern give a high response.

\subsection{Defining the Filter and Transformation Field}
Our strategy will be to operate on the gradients of both the pattern $P$
and the target image $I$.  In particular, we take the signal to be
\begin{equation}
H = \bigl\| \nabla I \bigr\|, \label{signal}
\end{equation}
and the transformation field $\T$ to be rotation by the angle $\theta$, where
\begin{align}
\theta_{\nabla I} = \textrm{atan2} \left( \frac{\partial I}{\partial y}, \ \frac{\partial I}{\partial x} \right) \label{theta}
\end{align}
and atan2 is the usual two-argument arctangent function.

To design the filter $F$, we consider what will happen during the
extended convolution when we place $F$ at some pixel $q$.  The values of
$F$ will be scattered, with weight proportional to the gradient magnitude
at $q$; in other words, the filter will have its greatest effect at edges
in the target image. Now, if $q$ were the only point with non-zero gradient
magnitude, the optimal filter $F$ would simply be the distribution that scatters
all of its mass to the single point $\widetilde{p}$ -- the pattern center relative to the
coordinate frame at $q$. When there are multiple points with non-zero
gradient magnitude, we set $F$ to be the ``consensus filter'', obtained
by taking the linear combination of the different distributions, with weights
given by the gradient magnitudes.

\begin{figure*}[!t] 
\centering
\includegraphics[width=.75\textwidth]{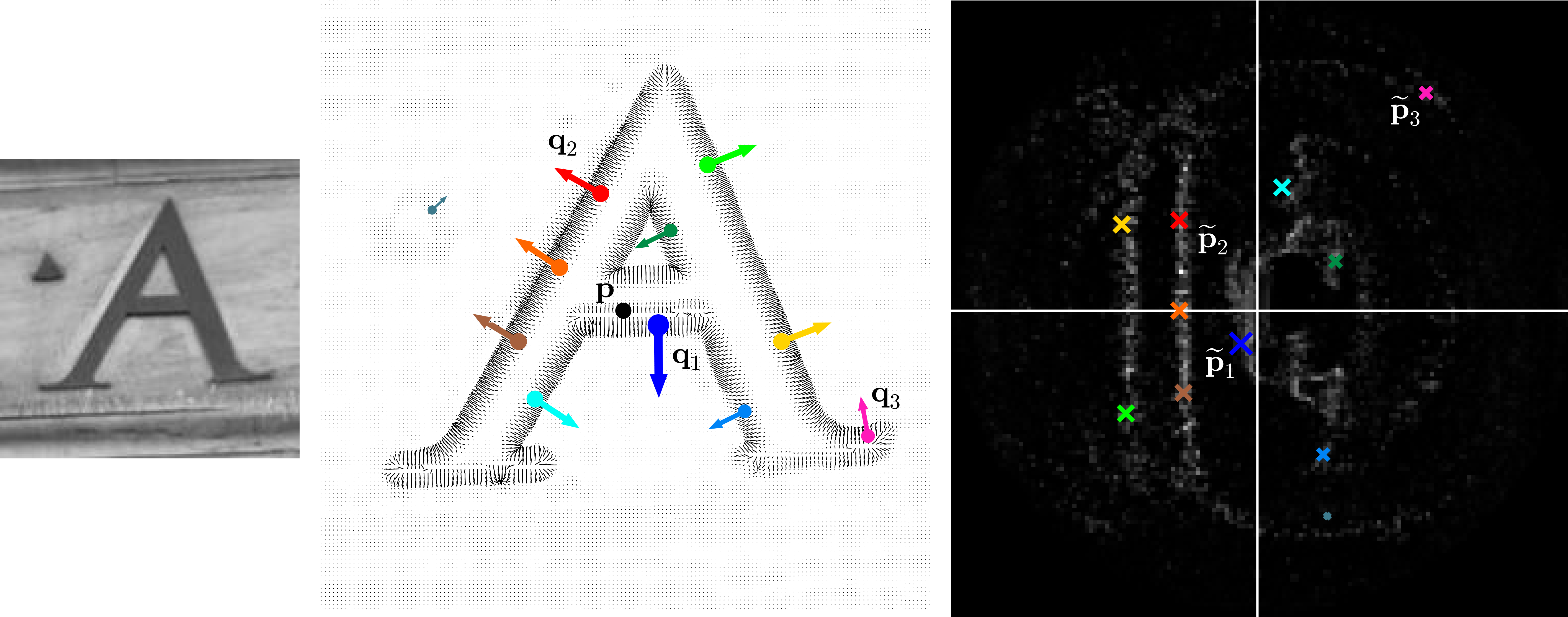}
\caption{Visualization of the construction of the optimal filter defined in Equation~(\ref{opt_filt}): A crop from an input image is shown on the left. The gradients, the keypoint $p$, and neighboring points $q_i$ are shown in the middle. The derived filter is shown on the right.}
\label{fig:ecd_construction}
\end{figure*} 

In practice, the filter is itself constructed by a voting operation. For example, consider Figure~\ref{fig:ecd_construction}, which shows an example of constructing the optimal filter (right) for an `A' pattern (left) with respect to its gradient field (middle) at the point $p$.  For each point $q$ in the vicinity of the pattern's center, the gradient determines both the position of the bin and the weight of the vote with which $q$ contributes to the filter. For example, since the gradient at $q_1$ is interpreted as the $x$-axis of a frame centered at $q_1$, the position of $p$ relative to this frame will have negative $x$- and $y$-coordinates. The gradient at $q_1$ has a large magnitude, so the point $q_1$ contributes a large vote to bin $\widetilde{p}_1$. The keypoint $p$ has positive $x$- and $y$-coordinates relative to the frame at $q_3$ but since the gradient is small, it contributes a lesser vote to bin $\widetilde{p}_3$.

Iterating over all points in the neighborhood of the pattern’s center, we obtain the filter shown on the right. While  the filter does not visually resemble the initial pattern, several properties of the pattern can be identified. For example, since the gradients along the outer left and right sides of the `A' tend to be outward facing, points on these edges cast votes into bins with negative $x$-coordinates, corresponding to the two vertical seams on the left side of the filter. Similarly, the gradients on the inner edges point inwards, producing the small wing-like structures on the right side of the filter.

Formally, we define the filter as:
\begin{equation}
F =\int \|\nabla P\| \ \rho_{-\T^{-1}(q) \cdot (p - q)} \ \delta \ dq,
\label{opt_filt}
\end{equation}
where the transformation field $\T$ is defined as rotation by the gradient
directions of the pattern $P$, and $\delta$ is the unit impulse, or Dirac
delta function.  This encapsulates the voting strategy described above.
In the appendix, we show that the filter $F$ defined in this
manner optimizes the response of the extended convolution at the origin.

\subsection{Discussion: Band-Limiting Revisited}
As we have seen, the extended convolution of an $N\times N$ image with
a rotating filter can be
computed in $O(N^3\log N)$ time by computing $O(N)$ standard
convolutions.  Though this is faster than the $O(N^4)$ brute force
approach, a similar form of pattern matching could be implemented in
$O(N^3\log N)$ by generating $O(N)$ rotations of the filter, performing
a convolution of the image with each one, and setting the value of the
response to be the maximum of the responses over all rotations.

\begin{figure}[hbt]
\centering
\epsfig{file=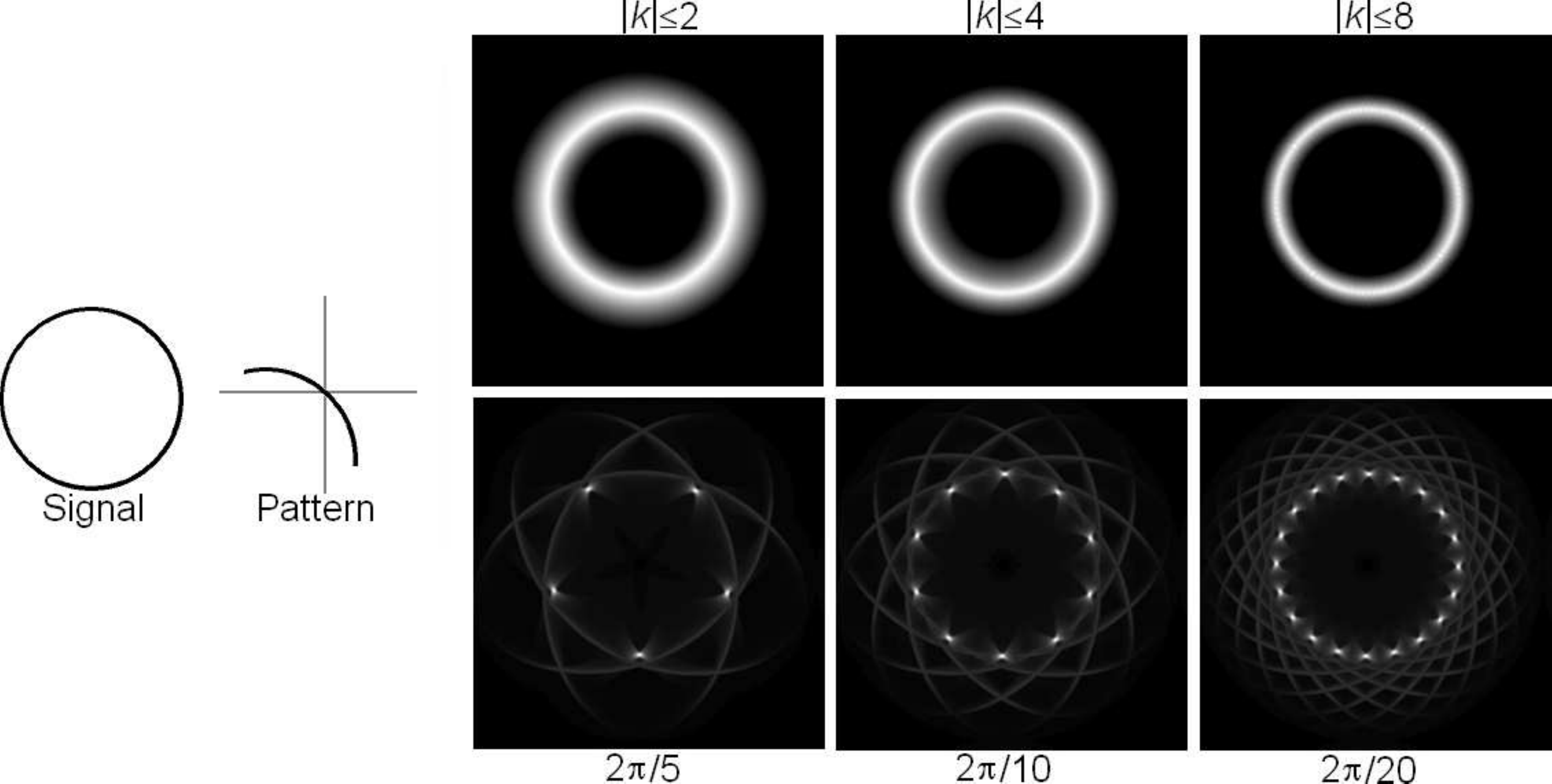,width=1.\columnwidth}
\caption{\footnotesize
\label{f:frequency}
Comparison of approximations to exact pattern detection.  Using subsets
of frequencies for extended convolution (top) converges more
quickly than convolution with multiple rotations of the pattern
(bottom).
}
\end{figure}

The difference becomes apparent when we consider limiting the number of
convolutions. As an example, Figure~\ref{f:frequency}, top, shows the
results of extended convolution-based pattern detection using low order
frequencies. The band-limiting in the angular component gives blurred
versions of the match-strength image, with the amount of blur reduced
as the number of convolutions is increased. In contrast, convolving the
image with multiple rotations of the pattern, as shown in the middle row,
yields sub-sampled approximations to the response image, and more standard
convolutions are required in order to reliably find all instances of the
pattern.  We can actually make a specific statement: the best way (in the
least-squares sense, averaged over all possible rotations) to approximate
the ideal extended convolution with a
specific number of standard convolutions is to use the ones corresponding
to the largest $a_i$: the most important projections onto the
rotational-Fourier basis.  Since in practice the lowest frequencies have
the highest coefficients, simply using the lowest few bases is a useful
heuristic.

\section{Application to Contour Matching}
\label{s:contour}
As a second test, we apply extended convolution to the problem of
matching complementary planar curves. To generate the signal and the
rotation field, we rasterize the contour lines and their normals into
a regular grid. We further convolve both the signal and the vector
field with a narrow Gaussian to extend the support of the functions.
Using these, we can define filters for queries and compute extended
convolutions.

Our contour matching algorithm differs from standard pattern matching in two ways. First, we are searching for complementary shapes,
not matching ones. Using the fact that complementary shapes
have similar local signals, but oppositely oriented gradient fields, we define the
filter using the negative of the query contour's gradient field.
Additionally, after finding the optimal aligning translation using
extended convolution, we perform an angular correlation to find the
rotation minimizing the $L_2$-difference between query and target.

\begin{figure}[tb]
\centering
\epsfig{file=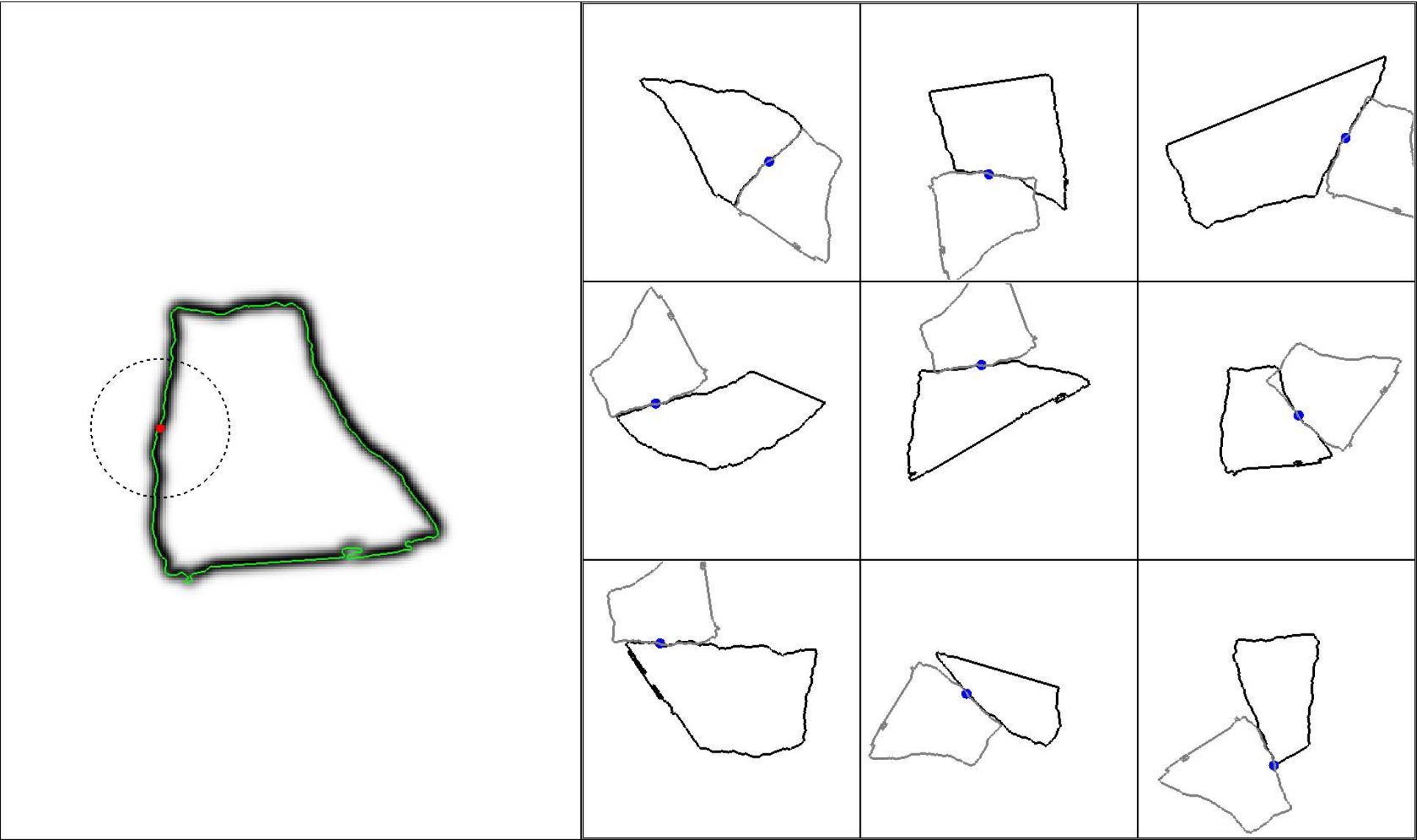,width=.85\columnwidth}
\caption{\footnotesize
\label{f:exampleSearch}
An example of applying the extended convolution to contour matching.
The image on the left shows the query contour with the region of
interest selected. The images on the right show the best nine
candidate matches returned by our system, in sorted order.
}
\end{figure}

An example search is shown in Figure~\ref{f:exampleSearch}.
The image on the left shows the query contour, with a black circle
indicating the region of interest. The image on the right shows the
top nine candidate matches returned using the extended convolution,
sorted by retrieval rank from left to right and top to bottom. Blue dots
show the best matching position as returned by the extended
convolution, and the complete transformation is visualized by
showing the registered position of the query in the coordinate
system of the target. Note that even for pairs of contours that do
not match, our algorithm still finds meaningful candidate alignments.

To evaluate our matching approach, we applied it to the contours of fragmented objects.  Reconstructing broken artifacts from a large collection of fragments is a labor-intensive task in archeology and other fields, and the problem has motivated recent research in pattern recognition and computer graphics~\cite{McBride:2003,Huang:2006,Brown:2008:ASF}.
As a basis for our experiments, we used the \emph{ceramic-3}
test dataset that is widely distributed by Leit\~{a}o and
Stolfi~\shortcite{Stolfi:TPAMI:2002}. This dataset consists of 112
two-dimensional fragment contours that were generated by fracturing five
ceramic tiles, and then digitizing the pieces on a flatbed scanner
and extracting their boundary outlines.

\begin{figure}[t]
\centering
\includegraphics[width=.9\columnwidth]{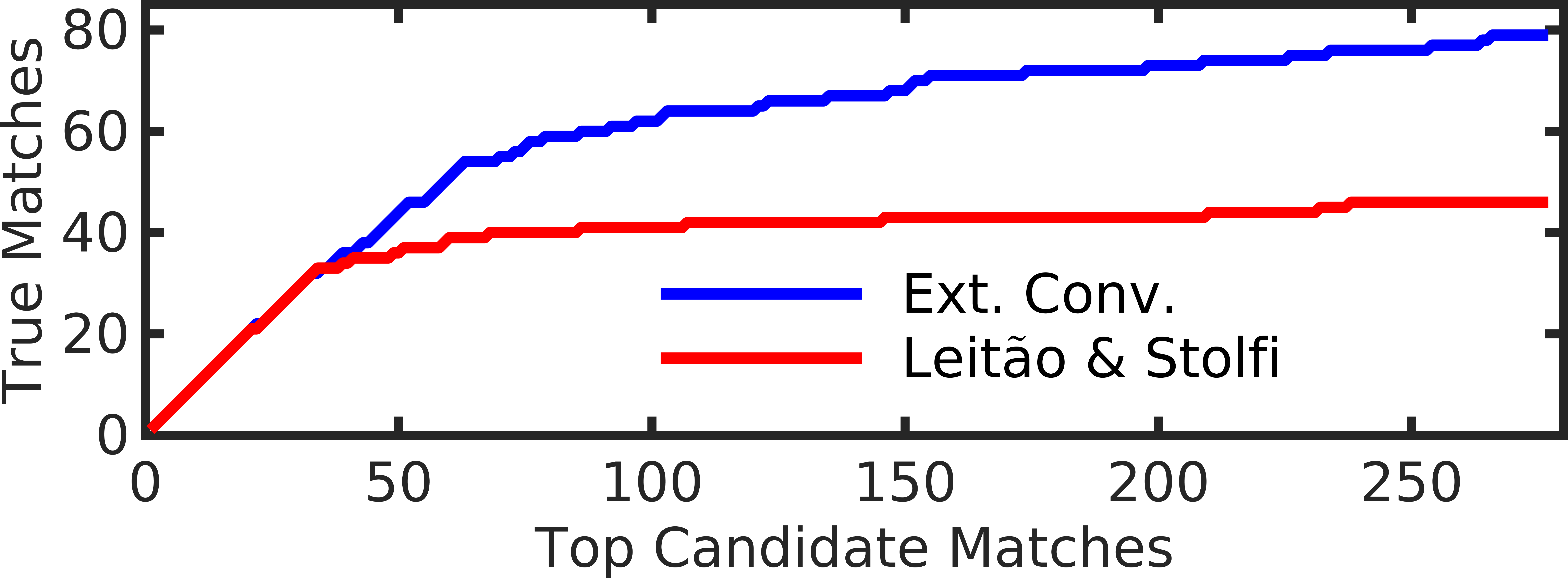}\caption{
The number of true contour matches within the first $n$ ranked candidate matches found using extended convolution, as compared to those found using method of Leit\~ao and Stolfi.}
\label{f:matching_results}
\end{figure}

Running the contour matching algorithm on each pair of fragments
produces a sorted list of the top candidate matching configurations for each fragment pair.  These candidate matches are reviewed to verify if they correspond to a true match. By using the same dataset,
we can directly compare our algorithm's performance against the
multiscale dynamic programming sequence matching algorithm and results
described in~\cite{Stolfi:TPAMI:2002}.  We used the same contour
sampling resolution as their finest multiresolution scale: fragments
thus ranged from 690 to 4660 samples per contour. The numbers of true matches found within the first $n$ ranked candidate matches found by the two algorithms
are compared in Figure~\ref{f:matching_results}.

The extended convolution matching algorithm outperforms the
multiscale sequence matching algorithm, and finds 72\% more correct
matches among the top-ranked 277 candidates. At this level of matching precision, our algorithm requires 6 hours to process the entire dataset of 112 fragments on a desktop PC (3.2 GHz Pentium 4). By reducing the sampling rate of the contour line rasterization grid or increasing the step size along the contours between extended convolution queries, the running time can be reduced significantly while trading off some search precision.  For collections with a large number of fragments, the matching algorithm can easily be executed in parallel on subsets of the fragment pairs.

\section{Application to Image Matching}
\label{s:evaluation}
An image feature descriptor can be constructed from the discretization of the optimal filter $F$, as defined in Equation~(\ref{opt_filt}), relative to the signal and frame field in Equations~(\ref{signal}) and~(\ref{theta}). We call this descriptor the \textit{Extended Convolution Descriptor} (ECD).

We compare the ECD image descriptor against SIFT in the context of feature matching on a challenging, large-scale dataset. We choose to compare against SIFT for several reasons. Foremost, SIFT has stood the test of time. Despite its introduction over two decades ago, SIFT is arguably the premier detection and description pipeline and remains widely used across a number of fields, including robotics, vision, and medical imaging.  Competing pipelines have generally emphasized computational efficiency and have yet to definitively outperform SIFT in terms of discriminative power and robustness\cite{karami2017image, tareen2018comparative}.

The advent of deep learning in imaging and vision has coincided with the introduction of a number of contemporaneous learned descriptors which have been shown to significantly outperform SIFT and other traditional methods in certain applications\cite{mishchuk2017working, he2018local, luo2018geodesc, Zhang:2019:LLD}. However, the performance of learned descriptors is often domain-dependent and ``deterministic" descriptors such as SIFT can provide either comparable or superior performance in specialized domains that learned descriptors are not specifically designed to handle\cite{Zhang:2019:HLU}. More generally, ``classical" methods for image alignment and 3D reconstruction, \textit{e.g.} SIFT + RANSAC, may still outperform state-of-the-art learned approaches with the proper settings\cite{schonberger2017comparative, jin2020IMW}. It is also worth noting that the extended convolution can be written as a convolution on \SE{2} (the group of planar rigid motions) by introducing Dirac delta functions to restrict contributions of integral over \SO{2} to those orientations defined by transformation field. Similar approaches have been shown to be an effective basis for image template matching \cite{kyatkin1999computation, kyatkin1999pattern, chirikjian2016harmonic}. 
    
The scope of this work is limited to local image descriptors -- we do not consider the related problem of feature detection. The SIFT pipeline integrates both feature detection and description in the sense that keypoints are chosen based on the distinctive potential of the surrounding area. As we seek to compare against the SIFT descriptor directly, we perform two sets of experiments. In the first, we replace the SIFT descriptor with ECD within the SIFT pipeline to compare practical effectiveness. The goal of the second experiment is to more directly evaluate our contribution with respect to the design of rotationally invariant descriptors. Specifically, we seek an answer to the following question: By having all points in the local region encode the keypoint relative to their own frames, do we produce a more robust and discriminating descriptor than one constructed relative to the keypoint's frame?

\subsection{Comparison Regime} \label{imRegime}
In both sets of experiments, we evaluate ECD and SIFT in the context of descriptor matching using the publicly available photo-tourism dataset associated with the 2020 CVPR Image Matching Workshop\cite{jin2020IMW}. The dataset consists of collections of images of international landmarks captured in a wide range of conditions using different devices. As such, we use the dataset to simultaneously evaluate descriptiveness and robustness. The dataset also includes 3D ground-truth information in the form of the camera poses and depth maps corresponding to each image.  In all of our experiments, we use the implementation of SIFT in the OpenCV library\cite{opencv_library} with the default parameters.

Due to the large size of the dataset, we restrict our evaluations to the image pools corresponding to six landmarks: \verb|reichstag|, \verb|pantheon_exterior|, \verb|sacre_coeur|, \verb|taj_mahal|, \verb|temple_nara_japan|, and \verb|westminster_abbey|, which we believe reflect the diversity of the dataset as a whole. Experiments are performed by evaluating the performance of the descriptors in matching a set of \textit{scene} images to a smaller set of \textit{models}.

For each landmark, five \textit{model} images are chosen and removed from the pool. These images are picked such that their subjects overlap but differ significantly in terms of viewpoint and image quality. The {\em scenes} are those images in the remainder of the pool that best match the models.

Specifically, SIFT keypoints are computed for all model in each pool. Keypoints without a valid depth measure are discarded. For each landmark, images in the pool are assigned a score based on the number of keypoints that are determined to correspond to at least one keypoint from the five models originally drawn from the pool.

Keypoints are considered to be in correspondence if the distance between their associated 3D points is less than a threshold value $\tau$. For each of the five models, all pixels with valid depth are projected into 3D using the ground-truth depth maps and camera poses.  These points are used to compute a rough triangulation corresponding to the surface of the landmark. As in \cite{guo2016comprehensive}, we define the threshold value relative to the area of the mesh, $A$,
\begin{gather}
    \tau = 0.005 \cdot \sqrt{ A \ / \ \pi \ }. \label{im_match_thresh}
\end{gather}
The top 15 images with the highest score from each pool are chosen as the \textit{scenes}. The scaling factor in the value of $\tau$ was determined empirically; it provides a good balance between keypoint distinctiveness and ensuring each scene contributes approximately 1000 keypoints to the total.

\subsection*{Comparisons within the SIFT Pipeline}
\begin{figure}[!t] 
\centering
\begin{subfigure}{\columnwidth}
\centering
\includegraphics[width=.9\columnwidth]{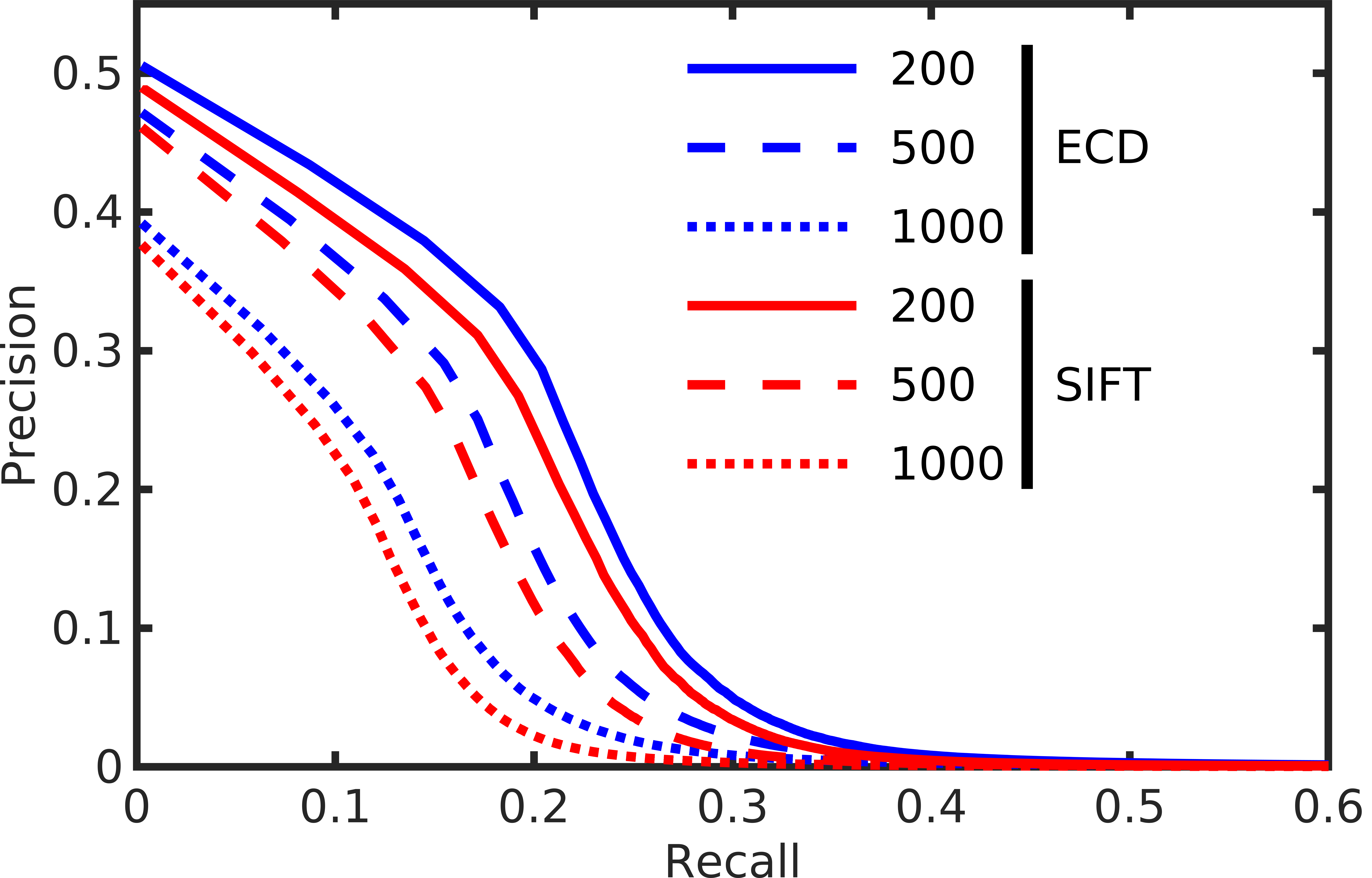}
\caption{Keypoints and scale determined by the SIFT feature detector}
\label{fig:image_pr}
\end{subfigure}
\\[3ex]
\begin{subfigure}{\columnwidth}
\centering
\includegraphics[width=.9\columnwidth]{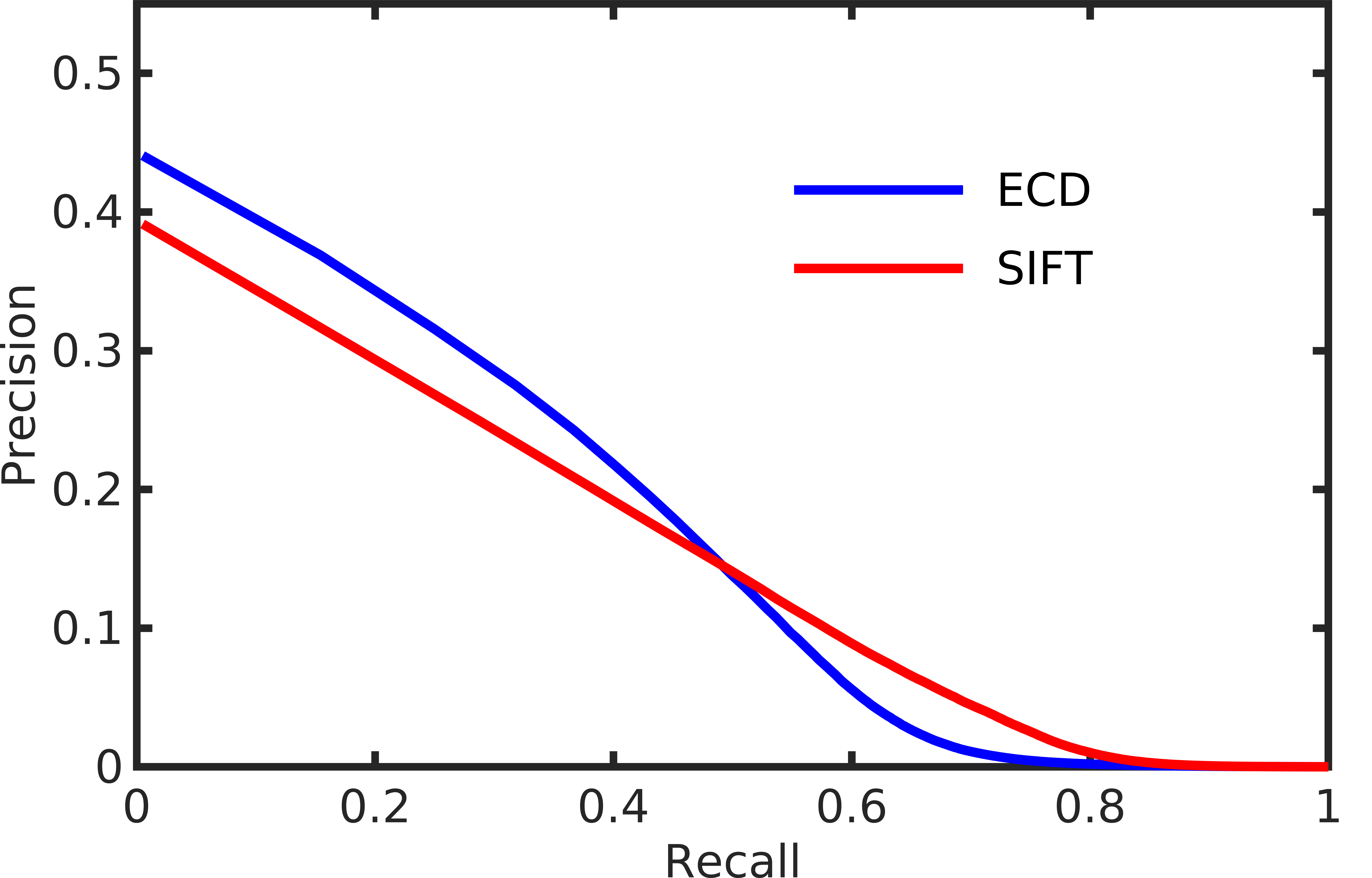}
\caption{Randomly selected keypoints and scale estimated from ground truth}
\label{fig:image_rand}
\end{subfigure}
\caption{The mean precision-recall curves for the ECD and SIFT descriptors. On the left, the keypoints and corresponding scales are computed in the SIFT pipeline. The three different curves correspond to the average over all scenes using the first 200, 500, and 1000 keypoints in each. On the right, keypoints are selected at random and scale is estimated from the ground truth. The curves are averaged over all scenes using 1000 keypoints in each. }
\label{fig:image_results}
\end{figure}

In our first experiment, we perform comparisons with keypoints selected using the SIFT keypoint detector to gauge ECD's practical effectiveness. For each model, we compute SIFT keypoints and sort them in descending order by ``contrast''\cite{lowe2004distinctive}. Of these, we retain the first 1000 distinct keypoints having a valid depth measure, preventing models with relatively large numbers of SIFT keypoints from having an outsize influence in our comparisons.  

For each scene, we compute SIFT keypoints and discard those without valid depth.  Those that remain are sorted by contrast and only the first 1000 distinct keypoints that match at least one keypoint from the five corresponding models are retained.

Next, ECD and SIFT descriptors are computed at each keypoint for both the models and the scenes. Both descriptors are computed at the location in the Gaussian pyramid assigned to the keypoint in the SIFT pipeline. The support radius, $\epsilon$, of the SIFT descriptor is determined by the scale associated with the keypoint in addition to the number of bins used in the histogram. The ECD descriptor uses more bins and we find that it generally exhibits better performance using a support radius 2.5 times larger that of the corresponding SIFT descriptor.

\subsection*{Comparisons using Randomized Keypoints}
Our second set of experiments are performed in the same manner using the same collection of models and scenes. The only difference is that the keypoints are selected at random so as to avoid the influence of the SIFT feature detection algorithm on the results. Specifically, for each model, 1000 keypoints are randomly chosen out of the collection of points that have a valid depth measure. Then, for each scene, we randomly select keypoints with valid depth and keep only those that correspond to at least one keypoint from the five associated images in the models.  This process is iterated until 1000 such points are obtained. 

We use the ground-truth 3D information to provide an idealized estimation of the scale. That is, for a keypoint, the associated 3D point is first translated  by $2 \tau$ in a direction perpendicular to the camera's view direction and then projected into the image plane. For both descriptors, the distance between the 2D keypoint and the projected offset defines the support radius.

\subsection*{Evaluating Matching Performance}
In both sets of experiments, we evaluate the matching performance of the SIFT and ECD descriptors by computing precision-recall curves for all keypoints in the scenes, an approach that has been demonstrated to be well-suited to this task\cite{ke2004pcasift, mikolajczyk2005GLOH}.  Given a scene keypoint, $s$ and corresponding descriptor $D(\vn{s})$,  all model keypoints are sorted based on the descriptor distance, giving $\{m_1,\ldots, m_M\}$ with
$$    \norm{ D(s) - D(m_i)} \leq \norm{ D(s) - D(m_{i+1})}.
$$
Some keypoints may be assigned multiple descriptors in the SIFT pipeline depending on the number of peaks in the local orientation histogram. In such cases we use the minimal distance over all of the keypoint's descriptors. 

Scene and model keypoints are considered to match if they correspond to the same landmark and the distance between their 3D positions is less than the threshold $\tau$ defined in Equation~(\ref{im_match_thresh}). We define $N_{s}$ to be the set of all model keypoints that are valid matches with $s$.  Following \cite{shilane2004princeton}, the precision $\mathcal{P}_{s}$ and recall $\mathcal{R}_{s}$ assigned to $s$ are defined as functions of the top $r$ model keypoints,
\begin{gather}
    \mathcal{P}_{s} (r) = \frac{ \abs{ N_{s} \cap \left\{m_i\right\}_{i \leq r}}}{r}
    \quad\hbox{and}\quad
    \mathcal{R}_{s} (r) = \frac{ \abs{ N_{s} \cap \left\{m_i\right\}_{i \leq r}}}{\abs{N_s}}.    \label{im_precision_recall}
\end{gather}

\begin{figure}[!t] 
\centering
\includegraphics[width=\columnwidth]{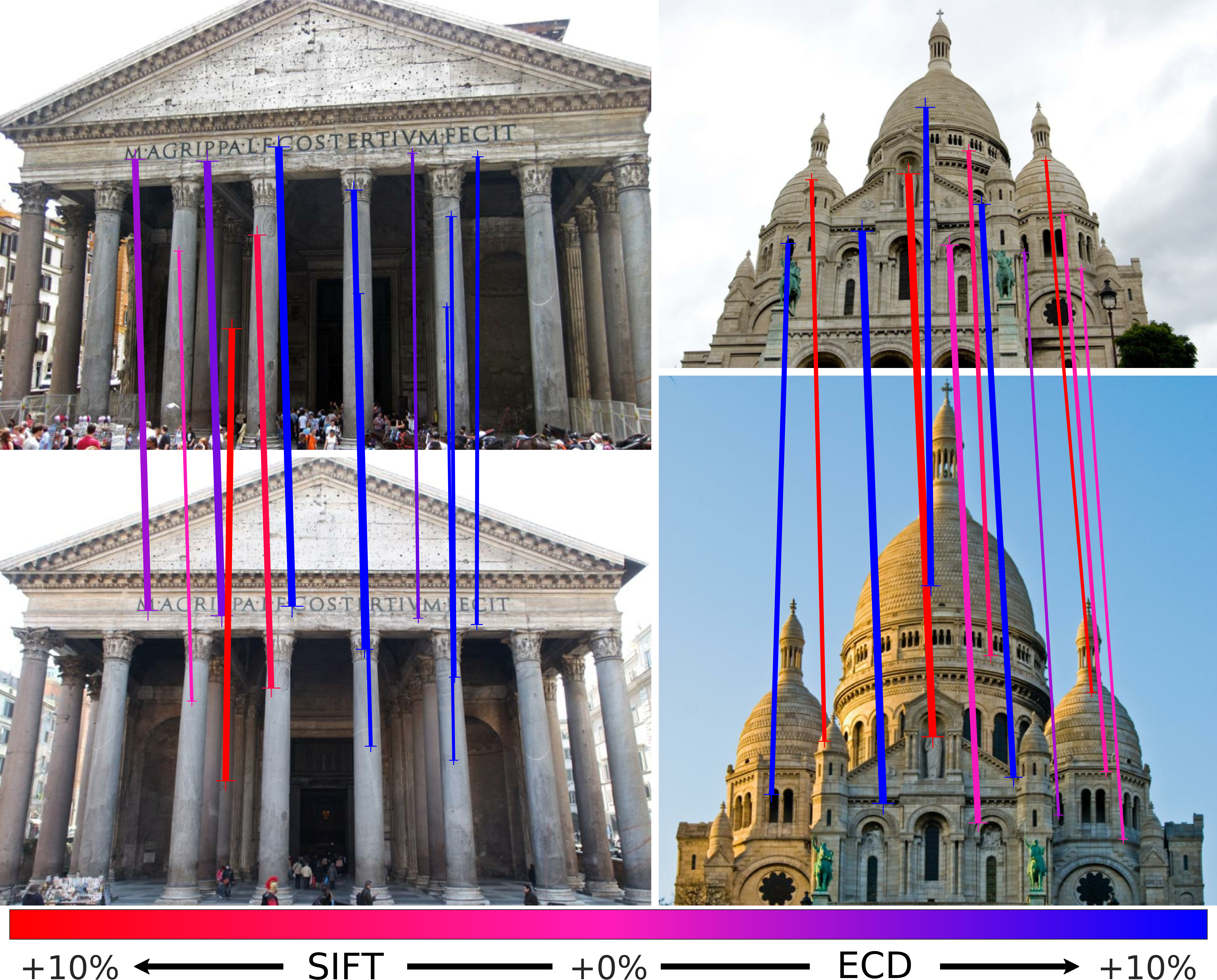}
\caption{Relative performance of SIFT and ECD in matching randomly selected keypoints in two pairs of scene (top) and model (bottom) images: Pairs of corresponding scene and model keypoints are grouped together and are visualized as vertical lines between the two images. Lines are colored to show the difference in the percentage of valid matches found by each descriptor and the thickness gives the number of corresponding pairs in each group.}
\label{fig:im_match_ex}
\end{figure}

\subsection{Results and Discussion}
We aggregate the results by computing the mean precision and recall across all keypoints in the scenes. For the first set of experiments, we compute three curves for each descriptor corresponding to the top 200, 500, and 1000 keypoints in each scene as ranked by contrast. The resulting precision-recall curves are shown in Figure \ref{fig:image_pr}. For the second set, we compute a single mean curve for each descriptor using all 1000 keypoints in each scene; these are shown in Figure \ref{fig:image_rand}.

Overall we see that ECD performs better than SIFT in our evaluations, though the difference is more pronounced when keypoint detection and scale estimation are decoupled from the SIFT pipeline as in our second set of experiments. In the former case, the precision of each descriptor decreases as the number of scene keypoints increases. This is not surprising as each successive keypoint added is of lower quality in terms of potential distinctiveness.  Figure \ref{fig:im_match_ex} shows a comparison of the valid matches found using the SIFT and ECD descriptors between two pairs of scene (top) and  model (bottom) images in the randomized keypoint paradigm.  We find that ECD tends to find slightly more valid matches than SIFT in less challenging scenarios, as in the case on the left where the scene and model image differ mainly in terms of a small change in the 3D position of the cameras. However, both descriptors perform similarly in more challenging scenarios as shown on the right. 

We do not argue that the results presented here show that the ECD descriptor is superior. Rather, they demonstrate that the ECD descriptor is distinctive, repeatable, and robust in its own right and has the potential to be an effective tool in challenging image matching scenarios.  However, it is important to note that effective implementations of the ECD descriptor may come at an increased cost. In our experiments, we find that ECD performs best with a descriptor radius of $7$, which translates to a descriptor size of $225$ elements, roughly two times the $128$ elements in the standard implementation of SIFT. 

The run-time of our proof-of-concept implementation of ECD does not compare favorably to the highly optimized implementation of SIFT in OpenCV. (SIFT runs up to a factor of ten times faster.) However, both approaches have the same complexity, requiring similar local voting operations to compute the descriptor, and we believe that ECD can be optimized in the future to be more competitive.

\section{Application to Image Filtering}
\label{s:smoothing}
\begin{figure}[hbt]
\centering
\epsfig{file=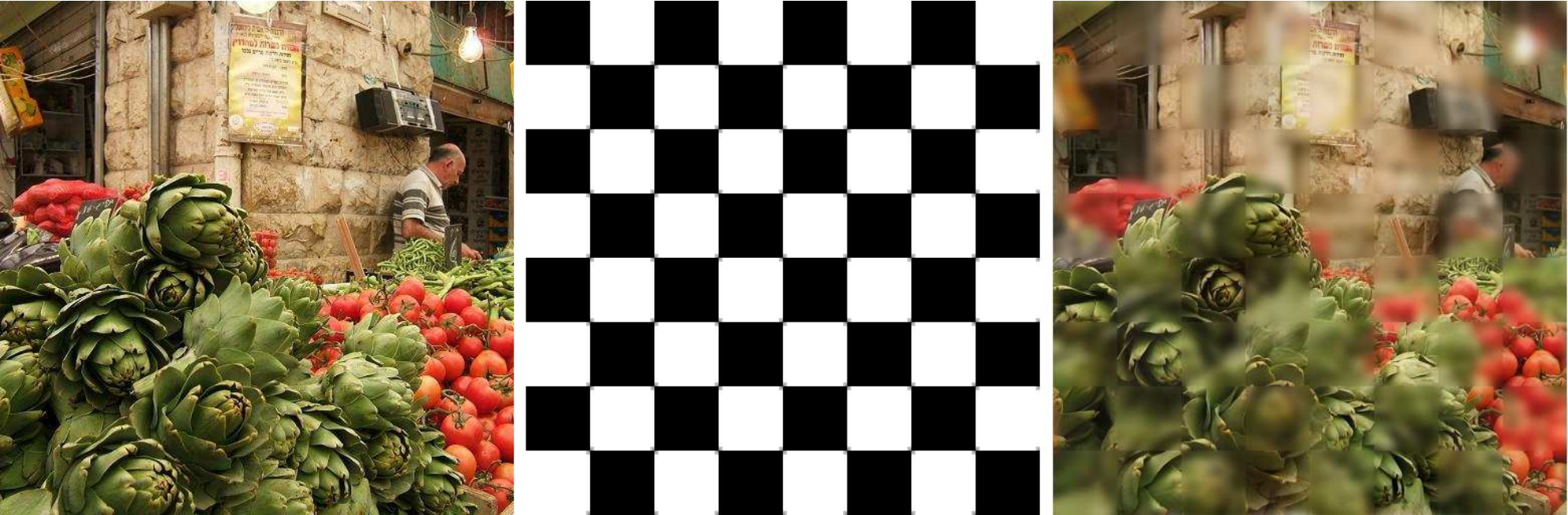,width=1.\columnwidth}
\caption{\footnotesize
\label{f:stall}
An example of using the extended convolution for adaptive smoothing. Given an image (left) and a transformation field
(center) the extended convolution can adaptively smooth the image (right) so that darker points in the transformation field maintain feature detail while lighter points are blurred out.
}
\end{figure}

We apply extended convolution to the problem of adaptive image
filtering, associating a scale or rotation to every pixel.
For example, Figure~\ref{f:stall} shows adaptive smoothing of a
market-stall image (left) with a Gaussian filter transformed according
to a checkerboard scaling mask (center).  White and black tiles in the
mask correspond to wide and narrow filters, respectively.

Guided by the principles outlined in section~\ref{s:representation}, we 
can decompose any filter into functions of the form
$$
F_k(r,\theta) = e^{ik\log r} f_k(\theta).
$$
Note the similarity to the case of rotation, with the Fourier transform
applied to $\log r$ instead of to $\theta$.  For the radially-symmetric
Gaussian filter, of course, $f_k(\theta)$ is just a constant and the
implementation becomes even simpler.

The result of
the extended convolution is shown at right, exhibiting the desired
smoothing effect with points overlapped by the white regions in the
transformation field blurred out and points overlapped by the dark
region retaining sharp details.

A similar technique is used in Figure~\ref{fig:teaser}~(right), but with
the scaling field obtained from the gradient magnitudes of the original
image.  As a result, the smoothing filter is scaled down at strong edges,
preserving the detail near the boundaries and smoothing away from them,
effectively acting similar to a bilateral filter~\cite{Tomasi:ICCV:1998,Slate:TOG:2006}.

To apply the extended convolution to image smoothing, we need
to modify the output of the extended convolution so that the value at
every point is defined as the weighted average of its neighbors. 
Treating the value $\bigl(\{H,\ \T\}*F\bigr)(p)$ as the weighted
sum of contributions from the neighbors of $p$, we can do this by dividing
the value at $p$ by the total sum of weights.  That is, if we denote by
$\{1,\ \T\}*F$ the extended convolution with a signal whose value is $1$
everywhere, the adaptively smoothed signal can be defined as:
\begin{equation}
\frac{\{H, \ \T\}*F}{\{1, \ \T\}*F}.
\label{eq:normalization}
\end{equation}

To localize the smoothing, we modify the
signal. Specifically, using the fact that scaling the filter $F$ by $\T(q)$ scales its integral by $\T^2(q)$, we normalize the signal $H$, setting:
$$\tilde{H}(q)=\frac{H(q)}{\|\T(q)\|^2}$$
so that the extended convolution $\{\tilde{H},\ \T\}*F$ distributes the
value $H(q)$ to its neighbors, using a unit-integral distribution.
Note that this modification is necessary only when the
transformation field $\T$ includes scaling; it is not needed when
$\T$ consists of rotations.

\begin{figure}[t]
\centering
\epsfig{file=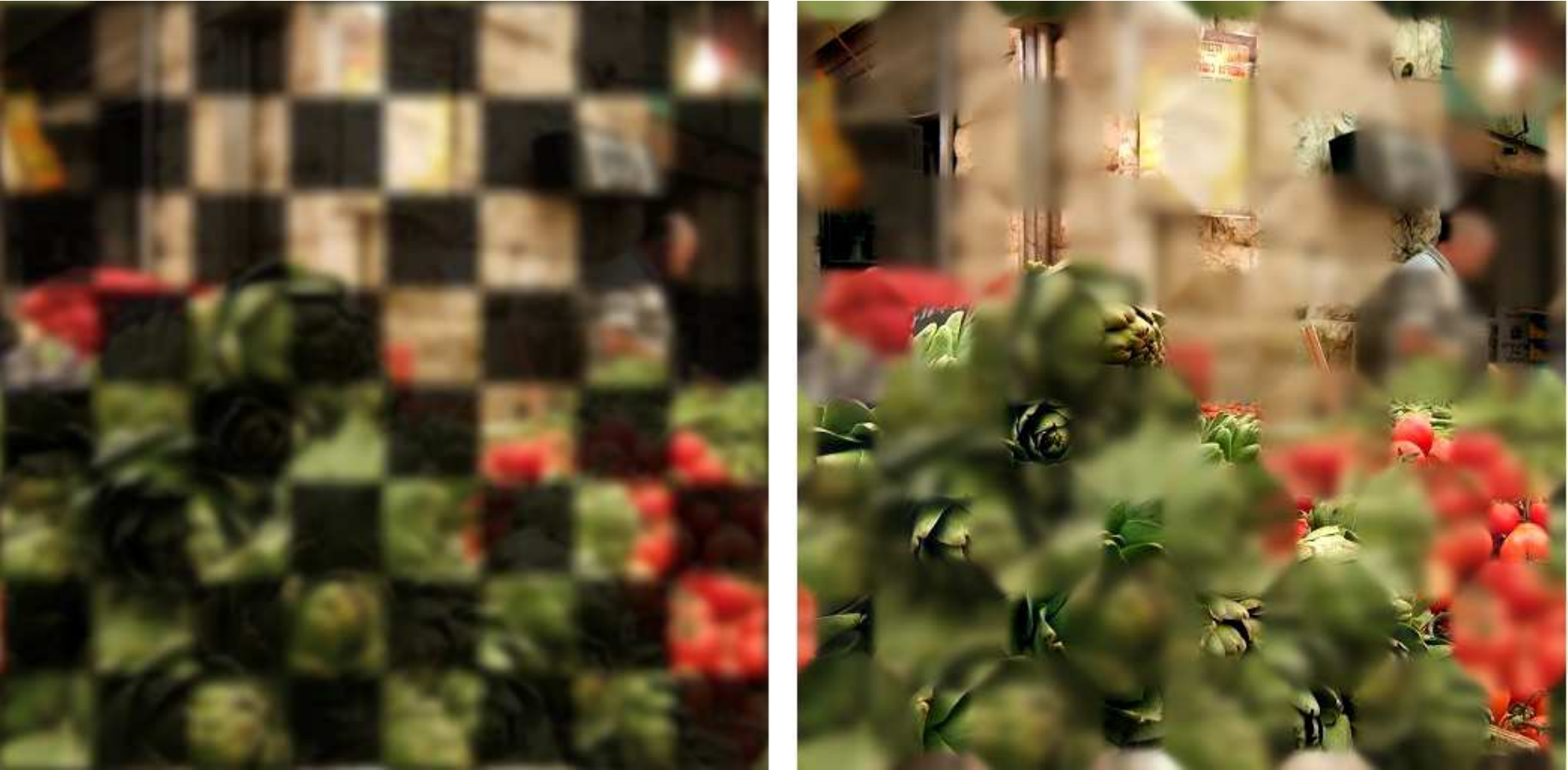,width=.7\columnwidth}
\caption{\footnotesize
\label{f:stall2}
\textbf{Left:} If the value of a pixel is not normalized by the
weighted average of its neighbors, the luminance is affected by the
filter variance at each pixel.  \textbf{Right:} Normalizing for filter
variance, but failing to account for filter scale change, results in
blur-bleeding across sharp edges in the transformation mask.}
\end{figure}

As an example, Figure~\ref{f:stall2}, left, shows the results of the
extended convolution for the market stall signal and checkerboard
transformation mask, without a division by $\{1,\ \T\}*F$.  Because the
filters in the black regions in the mask have smaller
variance, the corresponding regions in the image accumulate less
contribution and are darker.

Dividing by $\{1,\ \T\}*F$, we obtain Figure~\ref{f:stall2}, right.  The pixels
now have the correct luminance, but because the filters used in the
light portions are not normalized to have unit-integral, the adaptively
smoothed image exhibits blur-bleeding across the mask boundaries.
The correct result, with normalized $\tilde{H}$, is shown in
Figure~\ref{f:stall}, right.

\begin{figure}[b]
\centering
\epsfig{file=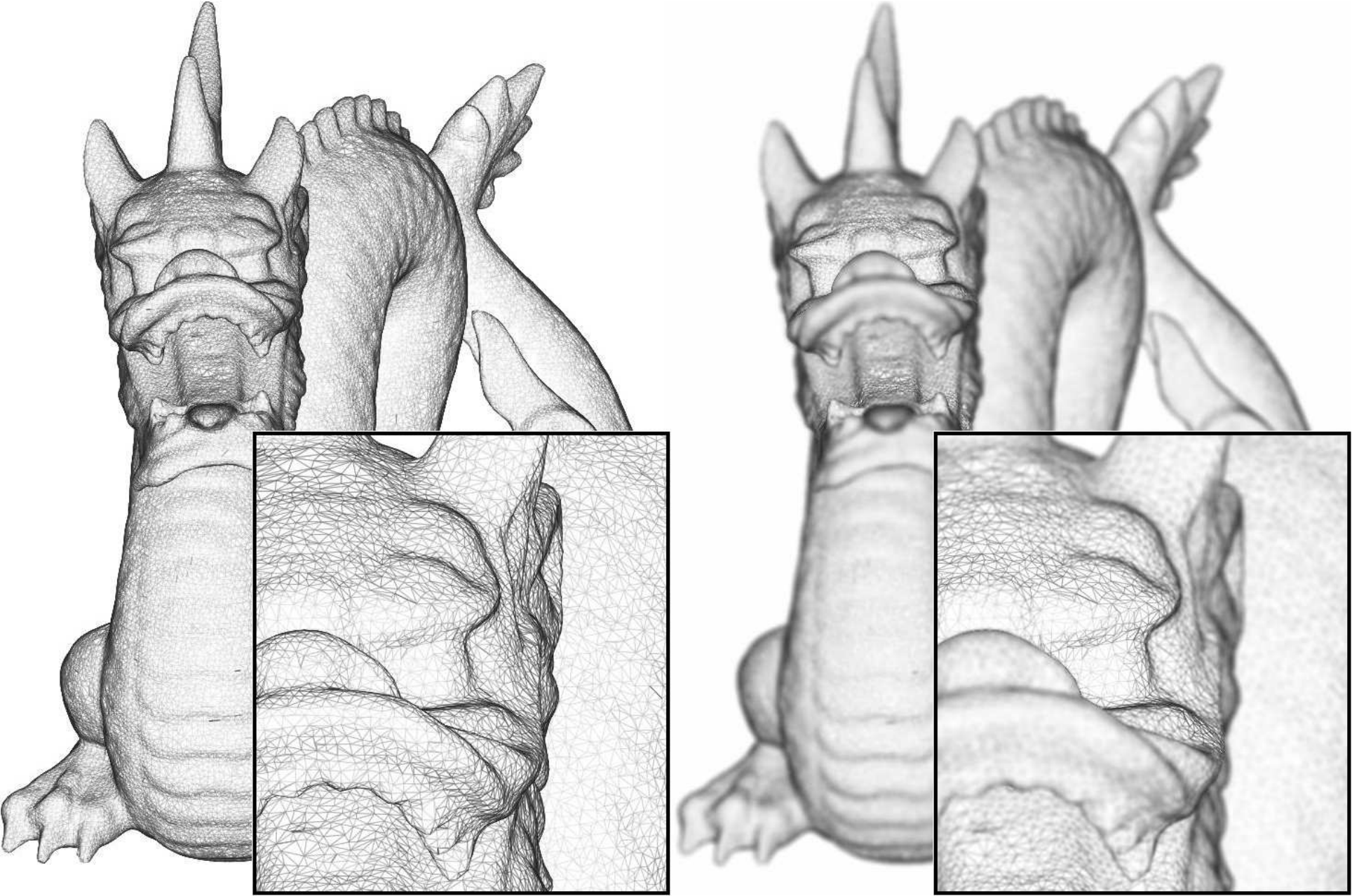,width=\columnwidth}
\caption{\footnotesize
\label{f:dragon}
\textbf{Left:} A wire-frame visualization of a dragon model.
\textbf{Right:} A simulation of depth-defocus obtained by
using the depth values to set the scaling mask in performing
adaptive smoothing on the wire-frame visualization.}
\end{figure}
An example of adaptive smoothing with a more complex scaling mask
is shown in Figure~\ref{f:dragon}. The image on the left shows a
wire-frame visualization of a dragon model and the image on the right
shows the results of adaptive smoothing applied to the visualization.
For the scaling mask, we set:
$$\T(p)=|Z(p)-Z(p_0)|$$
where $Z(p)$ is the value of the $z$-buffer at pixel $p$, and $p_0$
are the pixel coordinates of the center of the dragon's left eye. 
For the filter, we used the indicator function of a disk, smoothed
along the radial directions. Smoothing was necessary to ensure that
undesirable ringing artifacts did not arise when we approximated the
extended convolution by using only the first 64 frequencies. This
visualization simulates the depth-defocus
(e.g.~\cite{Potmesil:SIG:1981,Demers:GPUGems:2004,Scheuermann:SX3:2004,Kass:PIXAR:2006})
resulting from imaging the dragon with a wide-aperture camera whose
depth-of-field is set to the depth at the dragon's left eye.
Although the implementation does not take into account the depth-order of
pixels, and hence does not provide a physically accurate simulation of the
effects of depth-defocus, it generates convincing visualizations that can
be used to draw the viewer's eye to specific regions of interest.

The effectiveness of adaptive blurring is made possible by two properties:
First, despite the band-limiting of the filter, adaptive blurring accurately
reproduces fine detail, such as the single-pixel-width wire-frame lines in the
left eye. Second, because the extended convolution is implemented as a scattering
operation, it exhibits fewer of the edge-bleeding artifacts known to be difficult
(e.g.~\cite{Kass:PIXAR:2006}) in ``gathering'' implementations.

\begin{figure}[ht]
\newlength\imw
\imw=1.3in
\fboxrule=0.5pt
\fboxsep=0pt
\centering
\valign{\vfil\hbox{#}\vfil\cr
  \fbox{\includegraphics[width=0.5\imw]{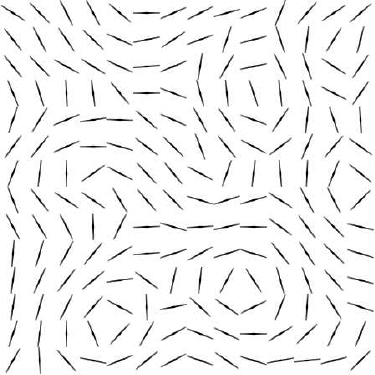}}\cr
 \,\fbox{\includegraphics[width=\imw]{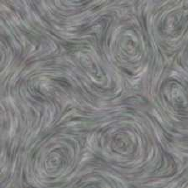}}\,\cr
  \fbox{\includegraphics[width=\imw]{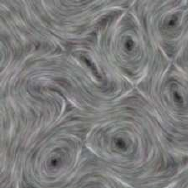}}\cr}
\caption{Line integral convolution for vector field visualization, implemented
via extended convolution of a random noise image with the rotational
field at left and a narrow anisotropic Gaussian filter.  At right, we
show the effects of not performing the normalization in
(\ref{eq:normalization})\,---\,while the result is not a correct
convolution, it is nevertheless an effective visualization.}
\label{fig:lic}
\end{figure}

As a further example of the effects achievable using adaptive filtering,
we demonstrate the use of extended convolution with a rotational field
to implement the Line Integral Convolution (LIC) technique for vector
field visualization~\cite{Cabral:SG:1993}.  We apply the extended
convolution of a long, narrow anisotropic Gaussian kernel to a random
noise image, using the given vector field's angle at each pixel to
determine the rotation to apply to the kernel.  The result is shown
in Figure~\ref{fig:lic}, center, while at right we show the result
produced when the normalization in (\ref{eq:normalization}) is not
performed.  The same technique was used to produce Figure~\ref{fig:teaser},
center: the gradient of the source image was used to define the rotational
field, and noise was added to the image before applying
extended convolution.

\section{Function Steering in 3D}
\label{s:steering3D}
One of the contributions of our presentation is that it allows function steering to be generalized to higher
dimensions. In this section, we discuss the limitations of using the classical formulation of function steering
to adaptively rotate filters in 3D, and describe how such filtering can still be supported within our generalized,
representation-theoretic framework.

As summarized in Section~\ref{ss:ClassicalSteering}, classical steerable filtering with a filter $F$ is
performed by using the functions $F_0,\ldots,F_{K-1}$ as a steering basis, where $F_j$ is the rotation of
the function $F$ by $j\pi/K$ and $K$ is the maximal angular frequency of the filter.

The efficiency of this implementation is based on three properties. (1)~The space spanned by the $F_j$
is the $K$-dimensional space containing the orbit of $F$, so the functions $F_j$ can be used to steer
the filter. (2)~The number of rotations, $K$, is equal to the dimension of the space spanned by the orbit,
so that the functions $F_j$ are the smallest set of functions required to steer $F$. And, (3)~the filter
is one of the basis functions, $F_0=F$, so that only $K$ of the functions $a_iF_j$ are non-zero, and
hence an implementation of extended correlation only requires $K$ standard correlations.

What limits the extension of this approach to 3D function steering is that it is impossible to generically
choose a set of $K$ rotations $R_0,\ldots,R_{K-1}\in\hbox{SO}(3)$ such that $R_0$ is the identity and the
functions $R_0(F),\ldots,R_K(F)$ are linearly independent. (See Appendix for more details.)

The inability to generalize classical steerable filtering to 3D has been
observed before, and it has been suggested that an expansion into spherical
harmonics might be used to accomplish this~\cite{Freeman:PAMI:1991}.  Our
generalized approach provides the details, showing how to compute the
extended correlation (resp.~convolution) in a manner analogous to the one
used for 2D rotations in Section~\ref{sec:rotation}.  In this discussion,
we will consider the spherical parameterization of the filter where we are
assuming that $K=O(n^{1/3})$ is the maximal angular frequency, so that
the dimensionality of each spherical function is $O(n^{2/3})$, and
the radial resolution is $N=O(n^{1/3})$.

\paragraph*{Filter Decomposition}
We first decompose $F$ as the sum of functions with differing angular frequencies:
$$
F = \sum_{l=0}^{K}\sum_{m=-l}^l F_l^m\qquad\hbox{with}\qquad F_l^m(r,\theta,\phi)=f_l^m\!(r)\,Y_l^m(\theta,\phi),
$$
where the functions $Y_l^m$ are spherical harmonics of frequency $l$ and index $m$.
This decomposition can be done by first expressing $F$ in spherical coordinates
and then running the Fast Spherical Harmonic Transform~\cite{SpharmonicKit} at each radius to get the
coefficients of the different frequency components.
\hfill [$O(n+n\log^2n)$]

\paragraph*{Standard Correlation}
Next, we compute the standard correlations of the signal with the functions $f_l^{m'}(r)Y_l^m(\theta,\phi)$:
$$G_l^{m,m'} = H \star f_l^{m'}F_l^m\qquad\forall\ l\in[0,K],\ m,m'\in[-l,l].$$
This can be done by first evaluating the function $f_l^{m'}F_l^m$ on a regular grid
and then using the 3D Fast Fourier Transform to perform the correlation.
\hfill [$O(n^2 + n^2\log n)$]

\paragraph*{Linear Combination}
Finally, we take the linear combination of the correlation results:
$$
\bigl(\{H,\ \T\}\star F\bigr)(p) =
 \sum_{l=0}^k\,\sum_{m,m'=-l}^l\overline{D_{m,m'}^l\bigl(\T(p)\bigr)}\,G_l^{m,m'}(p),
$$
where $D_{m,m'}^l:\hbox{SO}(3)\rightarrow{\mathbb C}$ are the Wigner-D functions, giving the
coefficient of the $(l,m')$-th spherical harmonic within a rotation of the $(l,m)$-th spherical
harmonic.
\hfill [$O(n^2)$]

Thus, our method provides a way for steering 3D functions, sampled on a regular grid with $n$ voxels, in
time complexity $O(n^2\log n)$. If, as in the 2D case, we assume that the angular frequency of the filter
is much smaller than the resolution of the voxel grid, $K\ll N$, the complexity becomes $O(nK^3\log n)$.


\section{Conclusion}
\label{s:conclusion}
We have presented a novel method for extending the convolution and
correlation operations, allowing for the efficient implementation of
adaptive filtering.  We have presented a general description of the
approach, using principles from representation theory to guide the
development of an efficient algorithm, and we discussed specific
applications of the new operations to challenges in pattern matching
and image processing.

In the future, we would like to apply extended convolutions using
transformation fields consisting of both rotations and isotropic scales. 
We believe that this type of implementation opens the possibility for
performing local shape-based matching over conformal parameterizations.

\section*{Acknowledgements}
This work was performed under National Science Foundation grant IIS-1619050 and Office of Naval Research Award N00014-17-1-2142.


\bibliographystyle{eg-alpha-doi} 
\bibliography{paper}

\appendix
\label{s:appendix}

\subsection*{Defining Optimal Filters}
Given an image $I$ and associated frame field $\T$ and signal $H$, we seek a filter $F$, supported within a disk of radius of $\varepsilon$, whose extended convolution with the image is maximized (up to scale) at a point $q_0$.

Expanding the expression for the evaluation of the extended convolution at $q_0$, using the fact that evaluation at a point can be expressed by integrating against a delta function ($\delta$) at that point, and switching the order of integration, gives
\begin{align*}
&\hspace{-.125in}\left( \{H, \ \T \} * F \right) (q_0) \\
&=\int H(q) \ F \left(\T^{-1}(q) \cdot (q_0-q)\right) \ dq \\
&=\int H(q) \ \left(\int F(p) \ \delta\left(p-\T^{-1} (q) \cdot (q_0-q)\right) \ dp\right) \ dq \\
&=\int F(p) \ \left(\int H(q) \ \delta\left(p-\T^{-1}(q) \cdot (q_0-q)\right)\ dq\right) \ dp \\
& = \int F (p) \ \left( \int H(q) \ \rho_{\T^{-1}(q) \cdot (q_0-q)} \left( \ \delta(p) \ \right) \ dq \right) \ dp.
\end{align*}

Thus, the filter $F$ supported within a disk of radius $\varepsilon$ that, up to scale, maximizes the response of the extended convolution at $q_0$, is
\begin{align}
\nonumber
F &= \int_{\|q_0-q\|\leq\varepsilon} H(q)) \ \rho_{\T^{-1}(q) \cdot (q_0-q)} \left( \delta  \right) \ dq\\
&= \int_{\|q\|\leq\varepsilon} H(q_0+q) \ \rho_{-\T^{-1}(q_0+q) \cdot (q)} \left(  \delta  \right) \ dq. \label{opt_filter2}
\end{align}

\subsection*{Function Steering in 3D}
To implement three-dimensional function steering with functions whose angular frequency is bounded by
$K$, we would need to choose $N=(K+1)^2$ rotations $R_0,\ldots,R_{N-1}\in\hbox{SO}(3)$ such that $R_0$
is the identity and the rotation of any (band-limited) filter $F$ could be expressed as the linear
combination of the rotations of $F$:
$$R(F) = \sum_{j=0}^{N-1}\alpha_j(R) R_jF$$
Here, $\alpha_j:\hbox{SO}(3)\rightarrow{\mathbb C}$ is the function giving the coefficients of the $j$-th
function and $\sum_{j=0}^K(2j+1)=(K+1)^2$ is the dimension of the space of spherical functions whose
angular frequency is bounded by $K$.

The problem is that such a choice of rotations and hence the definition of the coefficient functions $\alpha_j$,
needs to depend on the filter $F$. To see this, we show that for any choice of rotations, $R_1,\ldots,R_{N-1}$,
we can always find a spherical function $F$ whose orbit under the group of rotations spans an $N$-dimensional
space but has the property that the functions $R_0F,\ldots,F_{N-1}F$ are linearly dependent, and cannot
span the same space.

Consider the rotations $R_0$ and $R_1$, the former is the identity map, and the latter must be a rotation about some
axis, which (without loss of generality) we assume to be the $y$-axis. Consequently, any function that is axially symmetric
about the $y$-axis must be fixed by both rotations. In particular, this implies that any linear combination
of the zonal harmonics has to be fixed. On the one hand, this implies that functions $\{R_0F,\ldots,R_{N-1}F\}$
span a space whose dimension is no larger than $N-1$ (since $R_0F=R_1F$) on the other hand, we know that if the
coefficients of all the zonal harmonics are non-zero, the orbit of $F$ under the action of the rotation group must
span an $N$-dimensional space. Thus, it is impossible to express all rotations of $F$ using linear
combinations of $\{R_0F,\ldots,R_{N-1}F\}$.

Note that while this precludes the extension of classical steerable filtering to the steering of arbitrary functions
in 3D, a more restricted version can still be implemented if the space of filters is constrained. Such a restriction is described in the work
of Freeman and Adelson, where they discuss the possibility of filtering with functions that are rotationally symmetric
about the $y$-axis. Since the only rotations fixing such filters are rotations about the $y$-axis, this subspace
of functions may be steered if the rotations $R_1,\ldots,R_{N-1}$ do not fix the $y$-axis.

\end{document}